%% file: main.tex
\pdfoutput=1

\documentclass[11pt]{article}

\usepackage{emnlp2021}

\usepackage{times}
\usepackage{latexsym}
\usepackage{hyperref}
\usepackage{booktabs}
\usepackage{multirow}
\usepackage{amsmath}
\usepackage{amssymb}
\usepackage{graphicx}
\usepackage{xcolor}
\usepackage{pifont}
\usepackage{setspace}
\usepackage{scrextend}
\usepackage{arydshln}
\usepackage{hhline}
\usepackage{float}
\usepackage{tikz}
\usepackage{tikz-dependency}
\usepackage{xspace}
\usepackage{array}
\usepackage{paralist, tabularx}

\usepackage{enumitem}
\setlist{leftmargin=*}
\usepackage{array, makecell}
\usepackage{url}

\usepackage[T1]{fontenc}

\usepackage[utf8]{inputenc}

\usepackage{microtype}

\input{math_commands.tex}

\definecolor{darkgreen}{HTML}{228B22}

\newcommand{\graytext}[1]{\textcolor{lightgray}{\text{#1}}}
\newcommand{\graytextbf}[1]{\textcolor{lightgray}{\textbf{#1}}}

\newcommand{\cmark}{\text{\boldsymbol{\textcolor{darkgreen}{\ding{51}}}}\xspace}%
\newcommand{\xmark}{\text{\boldsymbol{\textcolor{red}{\ding{55}}}}\xspace}%

\newcommand{\roberta}{RoBERTa\xspace}

\newcommand{\wiki}{\textsc{Wikipedia}\xspace}

\newcommand{\bookcorpusAtEnd}{\text{{\normalsize B}{\footnotesize OOK}{\normalsize C}{\footnotesize ORPUS}}\xspace}
\newcommand{\bookcorpus}{\bookcorpusAtEnd}

\newcommand{\realnews}{\textsc{RealNews}\xspace}
\newcommand{\ccnews}{\textsc{CC-News}\xspace}
\newcommand{\storiesAtEnd}{\text{{\normalsize S}{\footnotesize TORIES}}}
\newcommand{\stories}{\storiesAtEnd\xspace}
\newcommand{\openwebAtEnd}{\text{{\normalsize O}{\footnotesize PEN}{\normalsize W}{\footnotesize EB}{\normalsize T}{\footnotesize EXT}}}
\newcommand{\openweb}{\openwebAtEnd\,}

\newcommand{\nelson}{\textsc{LKT}\xspace}

\newcommand{\blimp}{\textsc{BLiMP}\xspace}

\newcommand{\lama}{\textsc{LAMA}\xspace}

\newcommand{\cat}{\textsc{CAT}\xspace}

\newcommand{\olmpics}{\textsc{oLMpics}\xspace}

\title{Probing Across Time:  What Does RoBERTa Know and When?}

 \author{
 Leo Z.\ Liu$^{\clubsuit}$\thanks{\ \ Equal contribution.} \quad 
 Yizhong Wang$^{\clubsuit}$\footnotemark[1]  \quad 
 Jungo Kasai$^{\clubsuit}$ \\ 
 \bf Hannaneh Hajishirzi$^{\clubsuit}$$^{\heartsuit}$ \quad
 Noah A. Smith$^{\clubsuit\heartsuit}$ \\\\
  $^{\clubsuit}$Paul G. Allen School of Computer Science \& Engineering,\\ University of Washington, Seattle, WA, USA \\
  $^{\heartsuit}$Allen Institute for Artificial Intelligence, Seattle, WA, USA \\
 {\tt\{zeyuliu2,yizhongw,jkasai,hannaneh,nasmith\}@cs.washington.edu}}

\date{}

\begin{document}

\maketitle

\input{01_abstract}

\input{1_introduction}

\input{3_probing_across_time}

\input{4_learning_patterns}
\input{5_pretraining_comparison}
\input{6_downstream_finetuing}
\input{7_discussion}
\input{8_related}
\input{9_additional_discussion}
\input{10_conclusion}
\input{acknowledgement}

\bibliography{references}
\bibliographystyle{acl_natbib}
\newpage
\input{999_appendix}

\end{document}

%% file: math_commands.tex
\usepackage{amsmath,amsfonts,bm}

\def\eqref#1{equation~\ref{#1}}

\def\1{\bm{1}}

\DeclareMathAlphabet{\mathsfit}{\encodingdefault}{\sfdefault}{m}{sl}
\SetMathAlphabet{\mathsfit}{bold}{\encodingdefault}{\sfdefault}{bx}{n}



%% file: 01_abstract.tex
\begin{abstract}
 
Models of language trained on very large corpora have been demonstrated useful for natural language processing.
As fixed artifacts, they have become the object of intense study, with many researchers ``probing'' the extent to which they acquire and readily demonstrate linguistic abstractions, factual and commonsense knowledge, and reasoning abilities.
Recent work applied several probes to intermediate training stages to observe the developmental process of a large-scale model \cite{serious-idea-crash}.
Following this effort, we systematically answer a question: for various types of knowledge a language model learns, \emph{when} during (pre)training are they acquired? Using \roberta as a case study, we find:  linguistic knowledge is acquired fast, stably, and robustly across domains. Facts and commonsense are slower and more domain-sensitive. Reasoning abilities are, in general, not stably acquired. As new datasets, pretraining protocols, and probes emerge, we believe that probing-across-time analyses can help researchers understand the complex, intermingled learning that these models undergo and guide us toward more efficient approaches that accomplish necessary learning faster. 
\end{abstract}

%% file: 1_introduction.tex
\section{Introduction}
\label{sec:intro}

Current NLP approaches lean heavily on language models trained on very large corpora \cite{elmo,devlin_bert:_2018,gpt2,liu_roberta:_2019,gpt3}.  Many researchers have sought to interpret what kinds of knowledge are acquired during this ``pretraining'' phase
\cite{bert-look-at,visualizing-bert,revealing-secrets-bert,belinkov-etal-2020-interpretability-tutorial}.
Extending \citet{serious-idea-crash}, we systematically conduct probing across the pretraining iterations, to understand not just what is learned (as explored in numerous past analyses of fixed, already-trained models), but also \emph{when}. 
In this work, we aim to inform future work on more efficient pretraining (e.g., fewer iterations are needed to acquire some kinds of knowledge) and on understanding dependencies among different kinds of knowledge.

Specifically, we apply a \emph{probing across time} framework to the widely used \roberta masked language model \cite{liu_roberta:_2019}.  We reproduce the pretraining of \roberta and apply a  suite of probes at many checkpoint iterations across pretraining (\textsection\ref{sec:learning_patterns}).
Our rich probe suite covers a diverse range of desirable knowledge types: linguistic properties \cite{nelson-probe}, factual knowledge \cite{lama-probe}, and commonsense \cite{cats-probe} and basic reasoning capabilities \cite{olmpics-probe}.  Our main finding is that linguistic information tends to be acquired fast, factual and commonsense knowledge slower, and reasoning abilities are largely unlearned.

We next apply probing across time to instances of \roberta trained on text from varying domains and with varying amounts of data (\textsection\ref{sec:domain_and_size}). Our experiments show that the learning order and learning patterns of different types of knowledge generally hold regardless of the data variation. However, different data choices do have an impact on the learning speed and the final performance. Our findings suggest that the inclusion of data in more diverse domains is more important than the quantity alone. 


Finally, we compare probes across time with research benchmark task performance across time (\textsection\ref{sec:finetuning}). We find that most of these benchmark tasks (e.g., SST-2, \citealp{sst2-dataset}, and SQuAD, \citealp{squad1.1-dataset}) require a relatively small number of pretraining steps to achieve high performance, which is similar to the fast learning patterns shown by linguistic probes. Some other tasks that are designed to test more complex knowledge (e.g., ReCoRD, \citealp{ReCoRD} and WSC, \citealp{winograd-schema-challenge}) benefit from longer pretraining time, aligning well with our findings for the corresponding type of probes.

We expect that, as new pretrained models and new probes emerge, probing-across-time analyses can help synthesize evidence for models' capabilities.
We release our code, as well as all the pretraining checkpoints at \url{https://github.com/leo-liuzy/probe-across-time} to benefit future research. \


%% file: 3_probing_across_time.tex
\section{Probing Across Time}
\label{sec:probing_across_time}

\input{tables/tab_probe_examples_main_morelines}



The great success of pretrained language models has motivated researchers to characterize what kinds of knowledge they encode.  
Probing seeks to determine how much is known by the pretrained model, so its representations are used without finetuning.
Few or zero additional parameters are estimated to target the probing task so that success on the probing task is attributable to pretraining alone. 
%
Past probing work has applied probing analysis after pretraining is complete and compares \textit{different} models  (e.g., BERT vs.\ GloVe).
We extend probing to different parameter states (i.e., training checkpoints) over the \textit{same} model's pretraining trajectory.
By observing how probe performance changes over time, we hope to understand  not just what the pretrained language model knows, but also when.

To that end, we adopt a diverse set of probes aiming at different types of knowledge (\textsection \ref{sec:method:probes}) and apply those probes at different iterates of the estimated model across pretraining. 
This requires pretraining from scratch and checkpointing intermediate parameter estimates (\textsection \ref{sec:method:pretraining}).
Considering our computational budget, we choose \roberta \footnote{In our work, \roberta stands for \roberta-base.} as our case study because of its popularity in supporting downstream probes and tasks. We leave exploration of different objectives and architectures, e.g., GPT-2 \cite{gpt2}, as future work.
We also set up baselines as additional relative references to stand for the best and the worst expected results from pretraining (\textsection \ref{sec:method:baselines}).  See \textsection \ref{appendix:hyper} for implementation details and the computational cost.

\subsection{Probe Suite Construction}
\label{sec:method:probes}



\citet{belinkov-etal-2020-interpretability-tutorial} categorize existing probes into two families. \textit{Structural probes} train a lightweight classifier that predicts a label on top of the model's internal representations. Such probes are mostly used to test linguistic knowledge like parts of speech.
\textit{Behavioral probes}, on the other hand, do not rely on  additional parameters or training, but use the model as it is to make predictions.
For example, if a masked language model can predict ``Honolulu'' for the input ``Barack Obama was born in [MASK],'' we can conclude that the pretrained model learned Obama's birth place. 

We adopt a rich set of probing tasks from five existing and publicly available probing packages to systematically evaluate different types of encoded knowledge: \nelson and \blimp for linguistic knowledge, \lama for factual and commonsense knowledge, \cat for commonsense knowledge, and \olmpics for reasoning knowledge. 
See \textsection \ref{appendix:hyper} for links to those probes.
For clarity, we focus here on four or five probing tasks from each package and present the rest (which showed similar patterns) in \textsection \ref{appendix:reults}.
See Table \ref{tab:examples-main} for an overview of these probing tasks and examples. We briefly describe the goal and our setup of these five packages as follows.
\paragraph{\nelson}
\citet{nelson-probe} introduce a \textit{structural linguistic} probe suite for testing linguistic knowledge and transferability (\nelson) of contextual representations.
For all tasks in \nelson, we train a linear classifier model to predict the linguistic annotation of each word in a sentence.
Through the performance of the classifier, we measure how closely the encoded information in the word representations conforms to linguistic annotations from human experts.
Following \citet{nelson-probe}, we use learnable coefficients to weigh a sum of representations from all the transformer layers, and compute the input vector to the classifier. We measure the probe performance by accuracy or $F_1$ on the test sets. 

\paragraph{\blimp} 
\citet{salazar-etal-2020-masked} introduce a \textit{behavioral linguistic} probe suite on the benchmark of linguistic minimal pairs (\blimp, \citealp{blimp-probe}).
This benchmark isolates specific phenomena in syntax, morphology, or semantics such as island effects and subject-verb agreement.
As seen in Table \ref{tab:examples-main}, input sentence pairs differ only by a word or a short phrase, but contrast in grammatical acceptability.
We test whether \roberta scores the grammatical sentence higher than the ungrammatical one.
The score for a sentence is calculated by sequentially masking one word at a time and averaging the log probabilities of the masked words.
Since no additional parameters or training are involved, \blimp provides a complementary perspective to \nelson --- if probing \textit{without} training shows the same pattern as \textit{with} training, it strengthens our observation on linguistic knowledge.

\paragraph{\lama} 
\citet{lama-probe}
introduce a \textit{behavioral} probing package that tests \textit{factual} and \textit{commonsense} knowledge. Each example in \lama is a cloze-style question with its subject or object masked. 
By predicting the masked word with \roberta, we measure its ability to recover real-world facts. We only consider the examples whose masked words exist in the \roberta vocabulary and measure whether \roberta predicts the masked word with the highest probability.
\paragraph{\cat}

\citet{cats-probe} introduce \textit{behavioral commonsense} probes based on a series of existing commonsense datasets.
These probes measure whether the pretraining model can give a higher score to positive examples (sentences that align with commonsense) than negative examples (sentences that don't).
The score of each sentence is computed by sequentially masking one word at a time and averaging the log probabilities.
\paragraph{\olmpics}

\citet{olmpics-probe} introduce a \textit{behavioral} probe package that tests the model's \textit{reasoning} abilities including object comparison, taxonomy conjunction, and multi-hop composition. 
We adopt the multiple choice masked LM setup where the pretrained \roberta is required to fill in the mask by selecting words from 2--5 candidates. 
Different from other probing packages, \citet{olmpics-probe} show that pretrained LMs do not get a large improvement over baselines on most of the probing tasks. This suggests that these reasoning tasks present challenges for current pretrained models, but we still include this probe package because it offers tests with a different aim and thus different insights into pretraining \roberta.

\subsection{Baselines for Relative Performance}
\label{sec:method:baselines}

Probes are not a perfect, absolute measure of encoded knowledge. In particular, \citet{hewitt-probe-control} find that probing classifiers can memorize labeling decisions independently of the linguistic knowledge of the representations.
\citet{pimentel-probe-information} argue that 
a tighter estimate of the encoded knowledge can be obtained by complex probing models.
We ask whether targeted knowledge can be easily extracted with few or zero additional parameters (i.e., \textit{ease of extraction}, as suggested by \citealp{pimentel-probe-information}).
We treat the probing scores as relative performance; and we address their concerns by comparing \roberta with the the following baselines:
\begin{compactitem}
    \setlength\itemsep{-0em}
    \item \textbf{Random Guess} randomly selects one class label or token from the candidate pool.
    \item \textbf{Random Vector + Linear Classifier} uses a random vector to represent each type, and trains a linear classifier on the top to predict the label with the token vector being frozen. 
    \item \textbf{GloVe + Linear Classifier} uses GloVe vectors \cite{pennington-etal-2014-glove}, and trains a linear classifier on top to predict the label.
    \item \textbf{Original \roberta} probes the officially released checkpoint\footnote{\url{https://github.com/pytorch/fairseq/blob/master/examples/roberta/README.md}.} of \roberta base to see if our checkpoints are pretrained properly and can achieve reasonable performance.
\end{compactitem}
Moreover, our probing results on different checkpoints can illustrate the relative performance change during pretraining. 

\subsection{Pretraining Setups}
\label{sec:method:pretraining}

We choose base-size \roberta as a case study.
In order to conduct probing over time, we replicate the \roberta pretraining procedure and periodically save checkpoints for later probing.
To ensure that probe-task relevant text is uniformly distributed over batches, the entire data is shuffled before every epoch.
Our training setting follows closely the one prescribed in \citet{liu_roberta:_2019}, except for the following differences: \textbf{1)} we use the hyperparameter setting with 1M update steps and a reduced batch size of 256, instead of 125K steps and a batch size of 2,048;\footnote{We chose a smaller batch size for more fine-grained observations.} \textbf{2)} we use static masking \cite{devlin_bert:_2018} during data processing instead of dynamic masking to run the code on TPUs.
These differences can result in slightly worse performance on downstream tasks \cite{liu_roberta:_2019}.
However, due to the large data size, we believe it won't significantly change the learning patterns we found in later sections.

We save a checkpoint every 20K training steps and more frequently during the first 12,800 steps, resulting in about 62 checkpoints for each pretraining setting.\footnote{In a pilot study, we observed that the training and validation loss start to plateau at 50K training steps.} Then we probe all these checkpoints to estimate the knowledge encoded by the model at different training steps.


\paragraph{Pretraining data} 
The original \roberta was pretrained on \bookcorpus (4 GB, \citealp{bookcorpus-dataset}), English \wiki (12 GB), CC-News (76 GB, \citealp{ccnews-dataset}), \openweb (38 GB, \citealp{openwebtext-dataset}), and \stories (31 GB, \citealp{stories-dataset}). 
Since we do not have access to their version of filtered \ccnews, we use \realnews (120 GB, \citealp{realnews-dataset}) instead, which is similar, according to \citet{liu_roberta:_2019}. This difference in training data might partly explain the performance degradation from the original \roberta. All the other corpora remain the same. This leads to a total of 205 GB\footnote{We follow \citet{liu_roberta:_2019} to report data size by gigabytes of the uncompressed text in this paper. Our entire pretraining data contain 46 billion tokens after tokenization.} unprocessed text, and each training epoch 
makes 360,851 update steps (3 epochs in total for the 1M update steps).
In our later controlled experiments, we sample these corpora to compare domains and data sizes in \textsection \ref{sec:domain_and_size}.

%% file: tables/tab_probe_examples_main_morelines.tex
\begin{table*}[thb!]
\tiny
\setlength\tabcolsep{2.2pt}   
\def\arraystretch{1.3}      

\begin{tabular}{c|c|c|c|ll}
\toprule
\textbf{Package}            & \textbf{Knowledge}         & \textbf{Task}                                          & \textbf{Formulation  }                        & \multicolumn{2}{c}{\textbf{Examples}}                                                                                          \\
\cline{1-6}
\multirow{10}{*}{\begin{tabular}{@{}c@{}} \nelson  \end{tabular}}                        & \multirow{10}{*}{Linguistic}                   & \multirow{2}{*}{POS Tagging}                  & \multirow{6}{*}{Token Labeling}      & \multicolumn{2}{c}{\multirow{2}{*}{\begin{dependency}[style=above]
          \begin{deptext}[column sep=0.2em]
              \graytextbf{PRON} \& \graytextbf{AUX} \& \textbf{VERB} \& \graytextbf{ADV} \& \graytextbf{ADP} \& \graytextbf{DET} \& \graytextbf{NOUN} \& \graytextbf{PUNCT} \\
              \graytextbf{I} \& \graytextbf{'m} \& \textbf{staying} \& \graytextbf{away} \& \graytextbf{from} \& \graytextbf{the}  \& \graytextbf{stock}  \& \graytextbf{.} \\
          \end{deptext}
        \end{dependency}}}                                                                                                  \\
                                             &                                                &                                               &                                      & \multicolumn{2}{c}{}                                                                                                  \\
                                             \cdashline{3-3}[1pt/1pt]
                                             \cdashline{5-6}[1pt/1pt]
                                             &                                                & \multirow{2}{*}{Syntactic Chunking}           &      & \multicolumn{2}{c}{\multirow{2}{*}{
        \begin{dependency}[style=above]
            \begin{deptext}[column sep=0.1em]
              \graytextbf{B-NP} \& \graytextbf{B-VP} \& \graytextbf{B-PP} \& \textcolor{black}{\textbf{B-NP}} \& \textcolor{lightgray}{\textbf{I-NP}} \& \textcolor{lightgray}{\textbf{I-NP}} \& \graytextbf{O} \\
              \graytext{\textbf{Shearson}} \& \graytext{\textbf{works}} \& \graytext{\textbf{at}} \& \textcolor{black}{\textbf{American}} \& \textcolor{lightgray}{\textbf{Express}} \& \textcolor{lightgray}{\textbf{Co}} \& \graytext{.} \\
          \end{deptext}
        \end{dependency}}}                                                                                                  \\
                                             &                                                &                                               &                                      & \multicolumn{2}{c}{}                                                                                                  \\
                                             \cdashline{3-3}[1pt/1pt]
                                             \cdashline{5-6}[1pt/1pt]
                                             &                                                & \multirow{2}{*}{Name Entity Recognition}     &       & \multicolumn{2}{c}{\multirow{2}{*}{\begin{dependency}[style=above]
            \begin{deptext}[column sep=0.2em]
              \graytextbf{O} \& \graytextbf{O} \& \textcolor{lightgray}{\textbf{I-ORG}} \& \textcolor{black}{\textbf{I-ORG}} \& \textcolor{lightgray}{\textbf{I-ORG}} \& \graytextbf{O} \& \graytextbf{O} \& \graytextbf{O} \& \graytextbf{O} \\
              \graytextbf{By} \& \graytextbf{stumps} \& \textcolor{lightgray}{\textbf{Kent}} \& \textcolor{black}{\textbf{County}} \& \textcolor{lightgray}{\textbf{Club}} \& \graytextbf{had} \& \graytextbf{reached} \& \graytextbf{108} \& \graytextbf{.} \\
          \end{deptext}
        \end{dependency}}}                                                                                                  \\
                                             &                                                &                                               &                                      & \multicolumn{2}{c}{}                                                                                                  \\
                                             \cline{3-6}
                                             &                                                & \multirow{2}{*}{Syntactic Arc Predication} & \multirow{4}{*}{Token Pair Labeling} & \multicolumn{2}{c}{\multirow{2}{*}{\begin{dependency}[edge vertical padding=-0.4ex]
          \begin{deptext} 
              Peter \&[2em] and \& May \&[2em] bought \&[2em]  a \&[2em]  car \&[.0em] . \\
          \end{deptext}
          \deproot[edge unit distance=1.05ex, label style={text=gray,draw=gray}, edge style={gray,densely dotted,>=}, hide label]{4}{ROOT}
          \depedge[edge unit distance=1.1ex, label style={text=gray,draw=gray},  edge style={gray,densely dotted,>=}, hide label]{4}{2}{\tiny Subj}
          \depedge[edge unit distance=0.8ex, label style={text=gray,draw=gray}, edge style={gray,densely dotted,>=}, hide label]{2}{1}{\tiny Conj}
          \depedge[edge unit distance=0.8ex, label style={text=gray,draw=gray}, edge style={gray,densely dotted,>=}, hide label]{2}{3}{\tiny Conj}
          \depedge[edge unit distance=0.3ex, label style={text=gray,draw=gray}, edge style={gray,densely dotted,>=}, hide label]{6}{5}{\tiny Det}
          \depedge[edge unit distance=0.8ex, label style={text=red}, edge style={red!60!black,thick,>=}]{3}{4}{\tiny \xmark}
          \depedge[edge unit distance=1ex, label style={text=red}, edge style={red!60!black,thick,>=}]{1}{3}{\tiny \xmark}
          \depedge[edge unit distance=0.8ex, label style={text=darkgreen,}, edge style={darkgreen!60!black,thick,>=},]{4}{6}{\tiny \cmark} 
        \end{dependency}}}                                                                                                  \\
                                             &                                                &                                               &                                      & \multicolumn{2}{c}{}                                                                                                  \\
                                             \cdashline{3-3}[1pt/1pt]
                                             \cdashline{5-6}[1pt/1pt]
                                             &                                                & \multirow{2}{*}{Syntactic Arc Classification}     &  & 
                                             \multicolumn{2}{c}{
                                             \multirow{2}{*}{\begin{dependency}[edge vertical padding=-0.4ex]
          \begin{deptext} 
            Peter \&[2em] and \& May \&[2em] bought \&[2em]  a \&[2em]  car \&[.0em] . \\
          \end{deptext}
          \deproot[edge unit distance=1.05ex, label style={text=gray,draw=gray}, edge style={gray,densely dotted}, hide label]{4}{ROOT}
          \depedge[edge unit distance=1.1ex, label style={text=gray,draw=gray}, edge style={gray,densely dotted}]{4}{2}{\tiny Subj}
          \depedge[edge unit distance=0.8ex, label style={text=gray,draw=gray}, edge style={gray,densely dotted}]{2}{1}{\tiny Conj}
          \depedge[edge unit distance=0.8ex, label style={text=gray,draw=gray}, edge style={gray,densely dotted}]{2}{3}{\tiny Conj}
          \depedge[edge unit distance=0.8ex,]{4}{6}{\tiny Obj}
          \depedge[edge unit distance=0.3ex, label style={text=gray,draw=gray}, edge style={gray,densely dotted}]{6}{5}{\tiny Det}
        \end{dependency}}}                                                                                                  \\
                                             &                                                &                                               &                                      & \multicolumn{2}{c}{}                                                                                                  \\
                                             \cline{1-6}
\multirow{5}{*}{\blimp}                       & \multirow{5}{*}{Linguistic}                  & Irregular Forms                      & \multirow{5}{*}{\begin{tabular}{@{}c@{}} Comparing \\ Sentence Scores \\ \textbf{Expected:} \\
$\mathbb{S}( \cmark) >  \mathbb{S}(\xmark)$ \end{tabular}}                     &  \cmark Aaron \textit{\textbf{broke}} the unicycle.                                   & \xmark Aaron \textit{\textbf{broken}} the unicycle.                              \\
                                            \cdashline{3-3}[1pt/1pt]
                                             \cdashline{5-6}[1pt/1pt]
&                                                & Determiner-Noun Agree.                                    &                     &  \cmark Rachelle had bought that \textit{\textbf{chair}}                             & \xmark Rachelle had bought that \textit{\textbf{chairs}}.                        \\
                                             \cdashline{3-3}[1pt/1pt]
                                             \cdashline{5-6}[1pt/1pt]
                                             &                                                & Subject-Verb Agreement                                 &                      & \cmark These casseroles \textit{\textbf{disgust}} Kayla.                          & \xmark These casseroles \textit{\textbf{disgusts}} Kayla.                          \\
                                             
                                             \cdashline{3-3}[1pt/1pt]
                                             \cdashline{5-6}[1pt/1pt]
                                             &                                                & Island Effect                                            &                 & \cmark Which \textit{\textbf{bikes}} is John fixing?                   & \xmark Which is John fixing \textit{\textbf{bikes}}?                             \\
                                             \cdashline{3-3}[1pt/1pt]
                                             \cdashline{5-6}[1pt/1pt]
                                             &                                                & Filler Gap                                    &                      & \cmark Brett knew \textit{\textbf{what}} many waiters find.                          & \xmark Brett knew \textit{\textbf{that}} many waiters find.                      \\
                                             
                                             \cline{1-6}
\multirow{4}{*}{\lama}                        & \multirow{3}{*}{Factual}        & Google RE                                     & \multirow{4}{*}{\begin{tabular}{@{}c@{}}  Masked LM  \\ \textbf{Expected:} \\ $\forall w \in V_{\text{\roberta}} \setminus \{ \cmark \},$ \\ $\mathbb{P}( \cmark \mid \mathcal{C}) >  \mathbb{P}(w \mid \mathcal{C})$ \\ \end{tabular}}                 & \multicolumn{2}{l}{Albert Einstein was born in \textit{\textbf{{[}MASK{]}}}       \hfill  \cmark: \textit{\textbf{{[}MASK{]}}} $=$ 1879 }                            \\
                                             \cdashline{3-3}[1pt/1pt]
                                             \cdashline{5-6}[1pt/1pt]
                                             &                                                & T-REx                                         &                   & \multicolumn{2}{l}{Humphrey Cobb was a \textit{\textbf{{[}MASK{]}}} and novelist     \hfill  \cmark: \textit{\textbf{{[}MASK{]}}} $= \text{screenwriter}$ }                            \\
                                             \cdashline{3-3}[1pt/1pt]
                                             \cdashline{5-6}[1pt/1pt]
                                             &                                                & SQuAD                                         &                   & \multicolumn{2}{l}{A Turing machine handles \textit{\textbf{{[}MASK{]}}} on a strip of tape.  \hfill \cmark: \textit{\textbf{{[}MASK{]}}} $ = \text{symbols}$ }                       \\
                                             \cdashline{2-3}[1pt/1pt]
                                             \cdashline{5-6}[1pt/1pt]
                                             &   \multirow{1}{*}{Commonsense}       & ConceptNet                                    &                   & \multicolumn{2}{l}{You can use \textit{\textbf{{[}MASK{]}}} to bathe your dog.    \hfill \cmark: \textit{\textbf{{\textbf{[}MASK{]}}}} $= \text{shampoo}$ }                                \\
                                             \cline{1-6}
\multirow{11}{*}{\cat}                         & 
\multirow{11}{*}{Commonsense}                   & Conjunction Acceptability                     & \multirow{11}{*}{
\begin{tabular}{@{}c@{}} Comparing \\ Sentence Scores \\ \textbf{Expected:} \\
$\forall \xmark,$ \\ $\mathbb{S}(\cmark) >  \mathbb{S}( \xmark)$ \end{tabular}}                     
&  \cmark Jim yelled at Kevin  \textit{\textbf{because}} Jim was so upset. & \xmark Jim yelled at Kevin \textit{\textbf{and}}  Jim was so upset.                                                                                                  \\
                                            \cdashline{3-3}[1pt/1pt]
                                            \cdashline{5-6}[1pt/1pt]
                                             &                                                & Winograd                     &                     &  \cmark The fish ate the worm. The \textit{\textbf{fish}} was hungry.  & \xmark The fish ate the worm. The \textit{\textbf{worm}} was hungry.                                                                                                  \\
                                             \cdashline{3-3}[1pt/1pt]
                                             \cdashline{5-6}[1pt/1pt]
                                             &                                                & Sense Making                                  &              &        \cmark Money can be used for buying \textit{\textbf{cars}}. & \xmark Money can be used for buying \textit{\textbf{stars}}.                                                                                      \\
                                             \cdashline{3-3}[1pt/1pt]
                                             \cdashline{5-6}[1pt/1pt]
                                             &                                                & \multirow{4}{*}{SWAG}      &                      & \multicolumn{2}{l}{\cmark Someone unlocks the door and they go in. \textit{\textbf{Someone leads the way in}}.}\\
                                             &                                                &                &                       &    \multicolumn{2}{l}{\xmark Someone unlocks the door and they go in. \textit{\textbf{Someone opens the door and walks out}}.}\\
                                             &                                                &                &                       &    \multicolumn{2}{l}{\xmark Someone unlocks the door and they go in. \textit{\textbf{Someone walks out of the driveway}}. }\\
                                             &                                                &                &                       &    \multicolumn{2}{l}{\xmark Someone unlocks the door and they go in. \textit{\textbf{Someone walks next to someone and sits on a pew.}}}\\
                                             \cdashline{3-3}[1pt/1pt]
                                             \cdashline{5-6}[1pt/1pt]
                                             &                                                &  \multirow{4}{*}{Argument Reasoning}              &                     &   \multicolumn{2}{l}{
                                             \cmark People can choose not to use Google, \textit{\textbf{and since all other search engines re-direct to Google}}, }    \\
                                             &                                                &                 &                     &  \multicolumn{2}{l}{\;\;\;  Google is not a harmful monopoly.}      \\
                                             &                                                &                 &                     &   \multicolumn{2}{l}{
                                             \xmark People can choose not to use Google, \textit{\textbf{but since other search engines do not re-direct to Google}}, }    \\
                                             &                                                &                 &                     &  \multicolumn{2}{l}{\;\; Google is not a harmful monopoly.}      \\
                                             \cline{1-6}
\multirow{5}{*}{\olmpics} & \multirow{5}{*}{Reasoning} & Taxonomy Conjunction                          & \multirow{5}{*}{\begin{tabular}{@{}c@{}} Multiple Choice \\ Masked LM \\ \textbf{Expected:} $\forall \xmark,$ \\ $\mathbb{P}(\cmark \mid \mathcal{C}) >  \mathbb{P}(\xmark\mid \mathcal{C})$ \end{tabular}}                 & \multicolumn{2}{l}{A ferry and a floatplane are both a type of \textit{\textbf{{[}MASK{]}}}.  \hfill  \cmark vehicle  \xmark airplane \xmark boat}            \\
                        \cdashline{3-3}[1pt/1pt]
                        \cdashline{5-6}[1pt/1pt]
                         &                            & Antonym Negation                                &                  & \multicolumn{2}{l}{It was \textit{\textbf{{[}MASK{]}}} hot, it was really cold.  \hfill  \cmark not \xmark really}             \\
                         \cdashline{3-3}[1pt/1pt]
                        \cdashline{5-6}[1pt/1pt]
                         &                            & Object Comparison                             &                  & \multicolumn{2}{l}{The size of a airplane is usually much \textit{\textbf{{[}MASK{]}}} than the size of a house.  \hfill \xmark smaller \cmark larger}              \\
                         \cdashline{3-3}[1pt/1pt]
                        \cdashline{5-6}[1pt/1pt]
                         &                            & Always Never                                  &                  & \multicolumn{2}{l}{A chicken \textit{\textbf{{[}MASK{]}}} has horns.  \hfill   \cmark  never \xmark rarely  \xmark sometimes  \xmark often \xmark always}                \\
                         \cdashline{3-3}[1pt/1pt]
                        \cdashline{5-6}[1pt/1pt]
                         &                            & Multi-Hop Composition                         &                  & \multicolumn{2}{l}{When comparing a 23, a 38 and a 31 year old, the \textit{\textbf{{[}MASK{]}}} is oldest. \hfill  \cmark second \xmark first  \xmark third}
\\ 
\bottomrule
\end{tabular}
\caption{Representative tasks from selected probe packages. $\mathbb{S}(\cdot)$ scores a sentence by sequentially masking each word in the sentence and averaging the log probabilities.  $\mathcal{C}$ denotes the rest of the sentence, and $\mathbb{P}(\cdot \mid \mathcal{C})$ is the conditional probability distribution over the vocabulary given $\mathcal{C}$. $V_{\text{\roberta}}$ is the vocabulary of \roberta.}
\label{tab:examples-main}
\end{table*}

%% file: 4_learning_patterns.tex
\begin{figure*}[th] 
\centering
\includegraphics[scale=0.188]{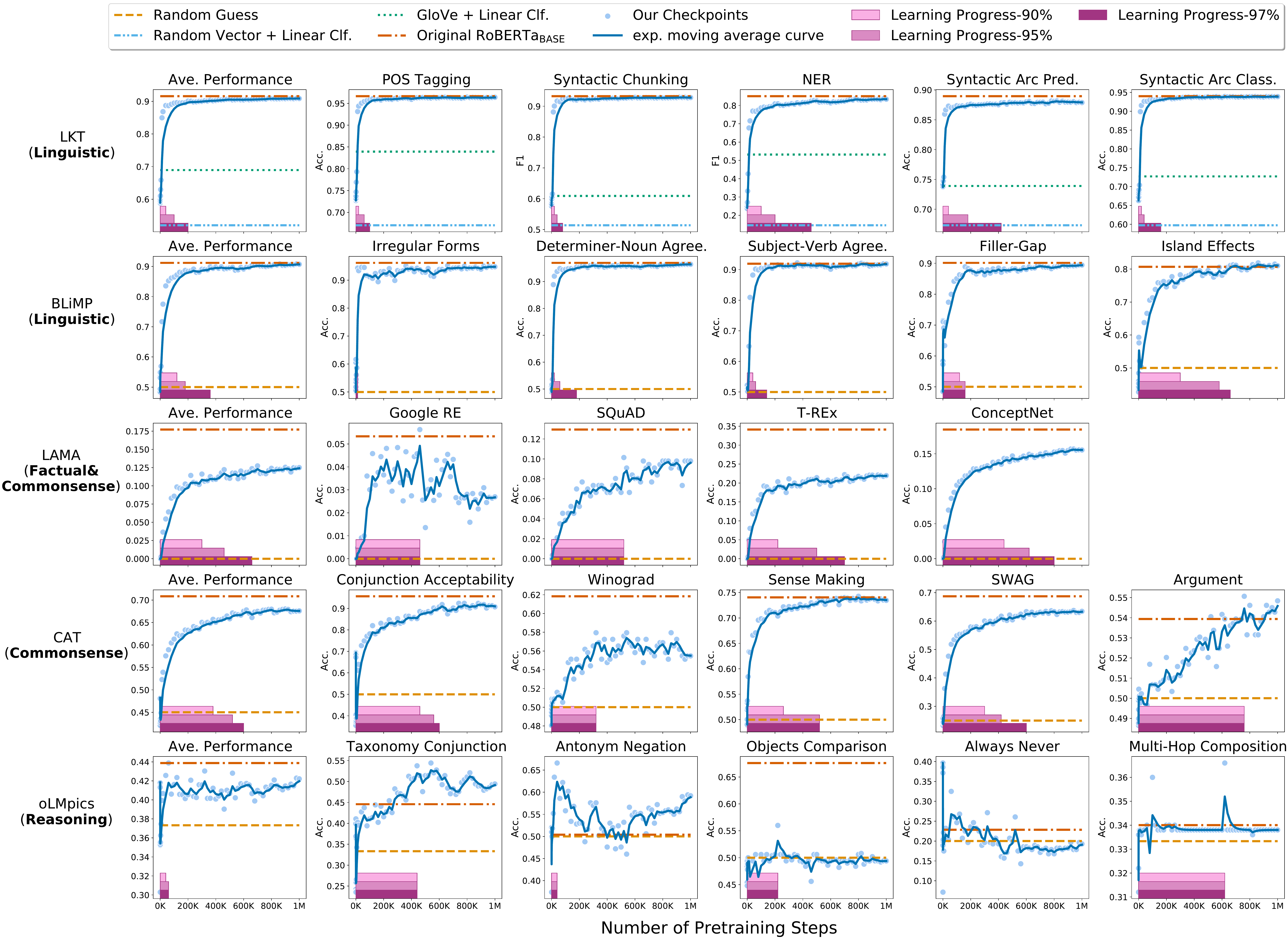}
\caption{Probes across time performance from our reproduced \roberta. Linguistic information tends to be acquired fast, factual and commonsense knowledge slower, and reasoning abilities are largely unlearned. For a better visualization, we use exponential moving average curve with a coefficient of $0.5$ to plot a smoothed curve.
For each probe package, the first column is the average performance over the 4-5 selected tasks. See complete plots in \textsection \ref{appendix:results:pattern}.  ``Learning Progress--$x$\%'' values are calculated with raw data, not smoothed data. The maximum performance of Always Never (\olmpics) occurs when the model is initialized. }
\label{fig:main}
\end{figure*}

\section{Learning Patterns}
\label{sec:learning_patterns}



In this section, we use our reproduced \roberta to ask and answer, \textbf{at which stage does the model acquire each kind of knowledge?} 
In addition to plotting probe performance across time (i.e., parameter updates during learning; see Fig.~\ref{fig:main}), a useful measurement is the number of updates required to reach $x$\% of the maximum performance achieved by our model across all iterations.  We denote this measurement  by ``Learning Progress--$x$\%'' for $x \in \{90,95,97\}$; it is indicated by the bottom horizontal bars in each probe's plot in Fig.~\ref{fig:main}.
We consider each type of knowledge in turn.
All of the following discussion is supported by Fig.\ \ref{fig:main}.

\subsection{Linguistic Learning}
\label{sec:learning_patterns:fast}

The structural linguistic probes test how closely the information in \roberta representations conforms to annotations developed from linguistic theories.
The behavioral probes test how sensitively the language model can respond to some detailed syntactic error.
In most cases, \roberta shows great success in learning linguistic knowledge with high speed and stability, and this pattern is consistent both in the classifier-based \nelson and behavioral \blimp probes (see the first two rows in Fig.\ \ref{fig:main}).
In the \nelson measurements from all 62 checkpoints, we observe that 97\% of the improvement in overall performance occurs within 20\% of the total training updates. 
The variation of this fast pattern among tasks in \nelson is small and all the performance converges closely to the originally reported results.
In a majority of \blimp tests, compared to \nelson, \roberta shows similar or even faster learning speed to achieve the 97\% threshold (Irregular Form, Determiner-Noun Agreement, and Subject-Verb Agreement), whereas some other tasks are more slowly learned.
See \textsection \ref{appendix:results:pattern} for more results.


\subsection{Factual and Commonsense Learning}
\label{sec:learning_patterns:slow}

Overall, Fig.~\ref{fig:main} shows \emph{slower} learning speed and more \emph{instability} in both \lama and \cat than in the linguistic probes. Most of our measurements require more than half of pretraining steps to achieve 97\% of the best performance. Compared with the high consistency in linguistic probes, there is more variation among tests. 
For example, the SQuAD and ConceptNet tests reveal the LM is steadily, although slowly, learning some factual and commonsense knowledge. However, other tests like argument reasoning and Winograd show  fluctuation or even a decrease in performance, and some factual knowledge such as Google RE is not easily learned.
We also note that there are noticeable gaps in the final performance of some tasks between our reproduced \roberta and original \roberta. We suspect that this is because of the several differences in our replication of \roberta (e.g., batch size and \realnews vs.\ \ccnews, \textsection \ref{sec:method:pretraining}). 

\subsection{Reasoning}
\label{sec:learning_patterns:non-learnable}

\olmpics shows some drastically different patterns from the other knowledge types (Fig.\ \ref{fig:main}). 
Many of the reasoning abilities are not learned during \roberta pretraining. For example, in Object Comparison, our \roberta model is on par with random guessing, and the performance on Always Never keeps decreasing overall after the initialization.
Some other tests included in \textsection \ref{appendix:results:pattern} also show severe fluctuation or similar patterns.
However, there are still signs of learning for some tasks. Taxonomy Conjunction shows the most promising sign of learning, and although to a small extent, \roberta also learns to do multi-hop reasoning. Another noteworthy observation is that \roberta's performance on Antonym Negation first increases rapidly, and then decreases in the rest of the first half of training; however, it starts to increase again toward the end. This indicates that this  knowledge is not stably stored in the model.

%% file: 5_pretraining_comparison.tex
\begin{figure*}[thb] 
\centering
\includegraphics[scale=0.252]{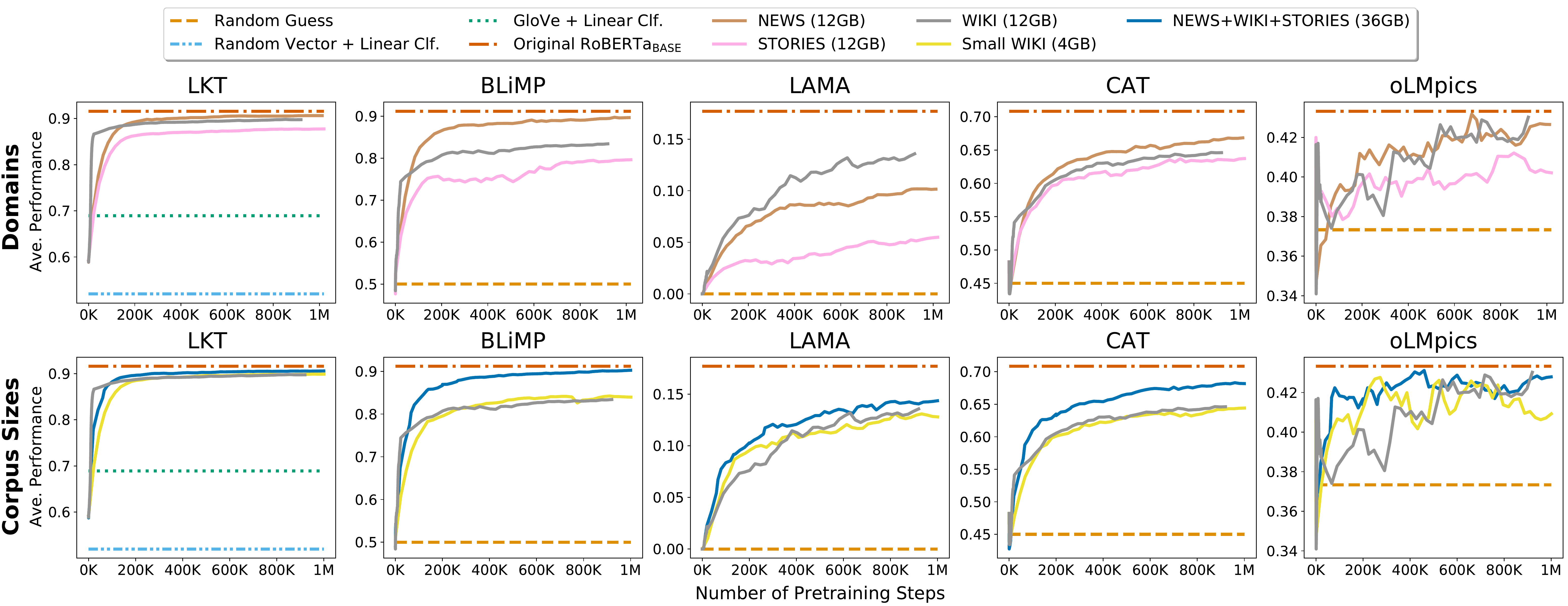}
\caption{Performance of probes across time, comparing pretraining domains and corpus sizes. This plot reveals that the learning order and patterns of different knowledge types generally hold regardless of the data variation, with higher impact from corpus domains. Performance for each probing package is averaged over its 4--5 selected tasks and smoothed with a coefficient of $0.5$ for better visualization (same as the first column in Fig.\ \ref{fig:main}).}
\label{fig:domains_and_sizes}
\end{figure*}

\section{Varying the Pretraining Corpus}
\label{sec:domain_and_size}


Our experiments on the pretraining trajectory in the previous section demonstrate that, when acquired, different types of knowledge are learned at different stages.
On the other hand, previous work \cite{raffel-etal-2020-t5,gururangan-etal-2020-dont} demonstrates that language domain and data size are important factors for a pretrained model's performance in downstream tasks.
Thus, the question arises: do our observations on the pretraining trajectory hold regardless of the training corpus?
In this section, we ask and answer \textbf{how do domains and corpus sizes affect the learning trajectory?}


\paragraph{Domains} With the same setting as \textsection \ref{sec:learning_patterns}, we pretrain \roberta with three controlled domains: English \wiki,
\realnews, 
and \stories.
We downsample \realnews and \stories to be roughly the same size as \wiki (12 GB).

The first row of Fig.\ \ref{fig:domains_and_sizes} compares probing performance across time over varying domains.
On one hand, the general learning pattern from the previous section persists regardless of the domain: linguistic knowledge is acquired faster than the other types.
On the other hand, we find that the change in the pretraining domain affects the final performance of all knowledge types but to different extents. \nelson is generally less affected than \blimp, probably due to the additional training of \nelson's classifier on the probing data. Although the model slowly acquires both factual and commonsense knowledge regardless of corpus domains, factual knowledge is much more affected by domain  (in fact, most affected among all).   \lama especially shows very slow learning on the \stories domain, implying that factual knowledge might be very sparse in \stories. The fact that the included factual tests---Google-RE, T-REx and SQuAD---are sourced from \wiki might explain why the factual tests show \roberta trained on \wiki learns faster than the one trained on \realnews. Though many existing works \citep{ROCStories, abdutive-commonsense-reasoning, back-to-the-future} used story data to study commonsense, pretraining the model on \stories still performs worst on our commonsesense probes (ConceptNet in \lama and most probes in \cat).
On the \olmpics probes, we keep observing large fluctuation on different domains, but do see more signs of learning on the \wiki and \realnews domains than on \stories. 
 See \textsection \ref{appendix:results:domain} for more detailed plots.
\paragraph{Corpus Sizes} To investigate the impact of corpus size, we experimented on downsampled English \wiki (4 GB), the original English \wiki (12 GB), and a combination of \wiki with downsampled \realnews and downsampled \stories (36 GB in total).

The second row of Fig.\ \ref{fig:domains_and_sizes} compares probing performance over varying corpus sizes.
In general, the learning order and learning patterns of different types of knowledge that we discuss in \textsection \ref{sec:learning_patterns} still hold for all data sizes. Comparing different data sizes, we find that the biggest corpus (with more diverse inclusion of domains) generally learns faster and results in an ultimately better \roberta in all tested knowledge. On the other hand, interestingly, comparing the downsampled English \wiki (4 GB) and original English \wiki (12 GB),  simply increasing the corpus size without changing domains does not improve the final probing performance substantially in all categories. This is even true for \lama, which tests factual knowledge relevant to \wiki. This suggests that diversity of data might be more important for pretraining than  quantity. See \textsection \ref{appendix:results:size} for more detailed plots.  



%% file: 6_downstream_finetuing.tex
\begin{figure*}[thb] 
\centering
\includegraphics[scale=0.168]{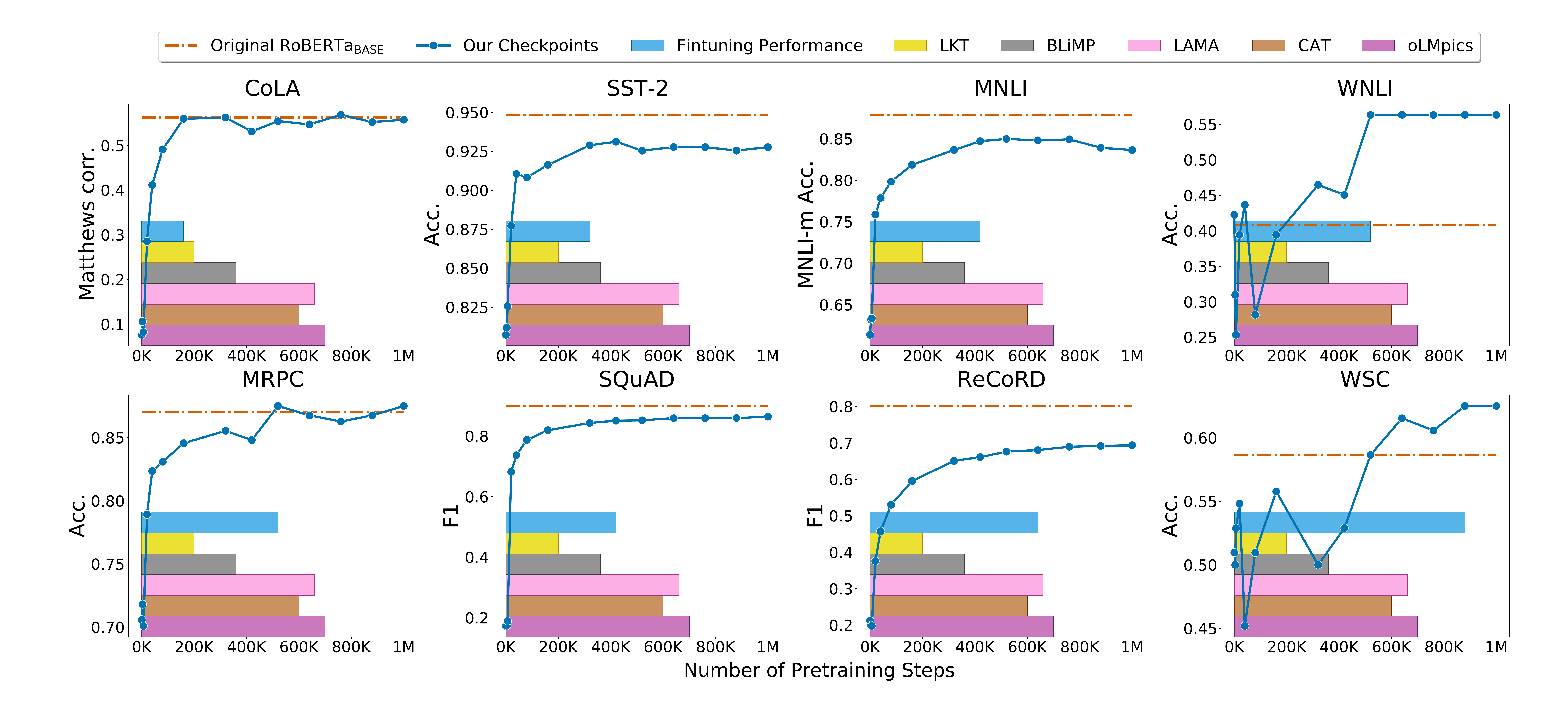}
\caption{Finetuning performance on research benchmarks using the checkpoints from our reproduced \roberta. Most of these benchmark tasks require less than half of pretraining steps to achieve high performance, while tasks that are designed to test more complex knowledge (e.g., ReCoRD and WSC) benefit from longer pretraining time. We plot Learning Progress--97\% of our \roberta in each probe package and finetuning (i.e., colored horizontal bars). The bottom five bars for the probes are shown for comparison. For Multi-genre NLI, we only show accuracy on the matched genres (same genres seen during finetuning), as we see similar curves on the unmatched genres.} 
\label{fig:finetune}
\end{figure*}

\section{Experiments on Research Benchmarks}
\label{sec:finetuning}

In \textsection \ref{sec:learning_patterns}-\ref{sec:domain_and_size}, we used probes to understand the knowledge learning process, but \textbf{what do those observations mean to more practical scenarios where people use pretrained \roberta with finetuning?} 
In this section, we provide insights into this question by finetuning our \roberta checkpoints on eight representative research benchmarks (e.g., SQuAD).
We conjecture that these benchmarks require more intermingled knowledge, whereas probing tasks usually target one specific phenomenon or type of knowledge.
By comparing them to the more controlled probes of \textsection \ref{sec:probing_across_time}, we aim to understand how acquisition of probed capabilities aligns with and perhaps accounts for performance on benchmarks. 


\paragraph{Experimental setup} We select 14 checkpoints from our pretraining to run our finetuning tasks: CoLA \citep{CoLA}, SST-2 \citep{sst2-dataset}, MNLI \cite{mnli-dataset}, WNLI (WSC recontructed as an inference task), 
MRPC \citep{MRPC}, 
SQuAD \citep{squad1.1-dataset}, ReCoRD \citep{ReCoRD}, and Winograd Schema Challenge (WSC, \citealt{winograd-schema-challenge}).
See Appendix \ref{appendix:hyper:finetuning} for hyperparameters.\footnote{To avoid impractical hyperparamter search for the large number of checkpoints we have, for each finetuning task, we use the same hyperparameters for all experiments. }

These tasks are chosen to reflect the diverse downstream use cases of pretrained language models, including single-sentence (CoLA, SST-2) or sentence-pair (MNLI, WNLI, MRPC) classification, question answering (SQuAD, ReCoRD), and multiple-choice classification (WSC).
Note that MNLI/WNLI and SQuAD/ReCoRD comprise two interesting contrastive pairs. Each pair shares a task format, with the latter tasks (i.e., WNLI and ReCoRD)  designed to rely more on commonsense than the former ones.  We expect more pretraining iterations for WNLI than MNLI and ReCoRD than SQuAD from the probing 
experiments (\textsection \ref{sec:learning_patterns}).

\paragraph{Results}
We plot our results in Fig.\ \ref{fig:finetune}. Different patterns are observed for different tasks.
CoLA and SST-2 require noticeably fewer pretraining steps, 
achieving 97\% of the best performance within 16\% and 32\% of the pretraining time, respectively, which is even faster than \nelson (20\%) or \blimp (36\%).
MNLI, WNLI, MRPC, and SQuAD are in a ``middle'' range (after linguistic knowledge, but before factual or commonsense knowledge), suggesting they are acquiring something the linguistic probes don't test but that is learned faster than what \roberta can learn for \lama~/~\cat. 
At last, ReCoRD and WSC are learned. In particular, WSC appears to be learned slower than all the tested probes, suggesting that pretraining is keeping learning knowledge that is beneficial to WSC but not tested in our probes.
Note again that ``learn the task'' is relative to the best performance we observe by our \roberta on the task.

As expected, WNLI requires longer pretraining than MNLI and ReCoRD longer than SQuAD, which aligns well with our finding in \textsection \ref{sec:learning_patterns:slow} that the model learns commonsense slowly as pretraining progresses. 
In addition, we observe MNLI performance even drops towards the end of pretraining, implying that longer pretraining does not necessarily lead to better finetuning performance. For interested readers, we also include a plot of correlation among all experimental results in \textsection \ref{appendix:correlation}

%% file: 8_related.tex
\section{Related Work and Further Discussion}
\label{sec:related}
\paragraph{Learning dynamics} Early work \cite{rumelhart} observed the dynamics in a feedforward neural network to assess the cognitive plausibility of a connectionist model. They found staged learning in past tense acquisition, similar to humans.
More recently, \citet{dynamics-by-svcca} studied linguistic and topic learning dynamics in hidden states of an LSTM language model.
They found that syntactic information is encoded at an early training stage, which is consistent with our finding despite the difference in training objective and network architecture.
Our work, instead, uses a rich set of probes to examine more diverse aspects of language and analyzes training iterations.

Concurrent work \citep{serious-idea-crash} is the closest work to ours and uses probes to investigate the learning dynamics as well. However, they find that linguistic knowledge and factual knowledge do not generally improve as pretraining proceeds, we find that factual and commonsense knowledge \emph{do} (\textsection \ref{sec:learning_patterns}); we attribute such difference to our more systematic choice of probes -- not only adding two more categories (i.e. commonsense and reasoning) but also more tasks in the linguistic and factual knowledge categories. For example, we found that the factual knowledge probed using SQuAD and ConceptNet data still increases as the pretraining progresses. However,  \citet{serious-idea-crash} only used a subset of T-REx, which plateaus quickly according to our experiments.
Sharing their concern of how data affects pretraining, we empirically 
investigate how the domain of pretraining corpus affects the dynamics of different types of knowledge.

\citet{idea-crash} investigate masked language models trained on corpora of varying sizes in a \textit{domain}.
They experiment with linguistic probes and show that 90\% of the improvement in syntactic probing performance can be achieved with a pretraining corpus of only about 10M words. In contrast, the probing performance in commonsense knowledge suffers from small training data. Different from the final models obtained in their work, we consider the \textit{entire pretraining trajectories}. Nevertheless, our findings by varying corpus size (\textsection \ref{sec:domain_and_size}) are consistent with their conclusion and additionally we find that adding more data in diverse domains can improve both types of knowledge. 
Their results on the relation between the corpus size and probing performance also support our finding that linguistic knowledge is generally easy to learn, while other types of knowledge require more efforts in term of both the data size and the training iterations. Since their experiments consist of different hyperparameter settings, and the domains for some of our datasets (i.e., \wiki and \stories) are the same as theirs, we consider our observation complementary to theirs.

\paragraph{Two views on probing} \citet{pimentel-probe-information} develop an \textit{information-theoretic} perspective that differs from our \textit{ease-of-extraction} view:  contextual representations can not have more information than the original sentence because the embedding function is deterministic. However, this view does not consider how they are \textit{structured} in the embedding space. These representations can only be taken advantage of if they are structured in an extractable way.
All the ``knowledge'' mentioned in our paper refers to such structured and easy-to-extract information.
In our pilot study, we experimented with MLP classifiers in structural probes.
They achieved much higher (almost highest in some cases) scores than the linear classifier on even the randomly-initialized \roberta model. 
Though agreeing with \citet{pimentel-probe-information}’s conclusion, MLP classifiers cannot reflect the learning process of the targeted knowledge and fail to explain why \roberta improves in downstream tasks over time (Fig. \ref{fig:finetune}).

%% file: 10_conclusion.tex
\section{Ethical Concerns}
\label{sec:ethical}
We estimate our carbon costs in Appendix \textsection \ref{appendix:hyper}. We recognize the possibility that changes to hyperparameters might lead to different conclusions and leave it to future work to balance the costs of such exploration with the value of a more detailed understanding of hyperparameter impact on learning over time.  We note that the robustness of learning patterns is partly supported by results from \citet{idea-crash}, since they probed models trained with different hyperparameters and observed similar linguistic patterns as we did.

\section{Conclusion}
We have shown how \textit{probing across time} reveals when, during pretraining iterations, a masked language model acquires various kinds of knowledge. \roberta, our case study model, is shown to learn linguistic knowledge faster than factual and commonsense knowledge, but  struggle to learn reasoning abilities.  We explored variation due to corpus domain and size, and related our findings to research benchmark tasks.  As models evolve and new probes emerge, we believe our \textit{probing across time} framework can serve as a general framework to inform progress on both fronts.

%% file: acknowledgement.tex
\subsubsection*{Acknowledgments}

In random order, we appreciate valuable discussion and feedback from Noah's ARK group at the University of Washington, Nelson Liu at Stanford University, Victoria Lin at Salesforce research,  Roy Schwartz at the Hebrew University of Jerusalem, and the anonymous reviewers. This research is supported in part by the Office of Naval Research under grants N00014-18-1-2826 and N00014-18-
1-2670.

%% file: 999_appendix.tex
\appendix

\section{Implementation and Hyperparameters}
\label{appendix:hyper}

\subsection{Pretraining}
\label{appendix:hyper:pretrain}
We chose the base-size \roberta model (125M parameters) as a case study for pretraining in this paper due to our computational budget. 

We used the TPU implementation\footnote{\url{https://cloud.google.com/tpu/docs/tutorials/roberta-pytorch}} of \roberta model in the official fairseq library.\footnote{\url{https://github.com/pytorch/fairseq}} 
However, because it needs to support experiments on TPUs, our implementation still has some differences compared to the original \roberta, which leads to the small performance gap between our pretrained models and the original \roberta baseline.
However, we believe that these differences won’t change the learning patterns very much. Specifically, for the static masking, since our data size is large, the whole pretraining (for reproducing \roberta) took less than 3 passes over the data. Therefore, using dynamic masking likely would not make much difference, especially considering that most of the probing performance becomes stable even before the first epoch ends.

The detailed pretraining hyperparameters are listed in Table \ref{tab:hyper:pretrain}.
Each pretraining was run using 8 TPU-v3 cores, and 1M steps took around 15 days. 
To increase awareness about the potential environmental impact of our large-scale pretraining, we use a tool from \citet{green-algo-estimator}\footnote{\url{http://www.green-algorithms.org/}} to estimate the energy and carbon cost of our experiments.
As a rough estimate, each pretraining (on one corpus) consumes about 684.02 kWh energy and has 173.19 kg CO2e carbon footprint. Therefore, in total, our pretraining experiments consume 4104.12 kWh energy and have 1039.14 kg CO2e carbon footprint.




\subsection{Probing and Finetuning}
\label{appendix:hyper:finetuning}
\paragraph{Probing} We ran our probing on checkpoints reported in Table \ref{tab:num-checkpoints}. All the probing packages used in our paper are publicly available. We use \nelson \cite{nelson-probe}\footnote{\url{https://github.com/nelson-liu/contextual-repr-analysis}} and \blimp \cite{blimp-probe}\footnote{\url{https://github.com/awslabs/mlm-scoring}} for probing linguistic knowledge; \lama \cite{lama-probe}\footnote{\url{https://cloud.google.com/tpu/docs/tutorials/roberta-pytorch}} for factual and commonsense knowledge; \cat \cite{cats-probe}\footnote{\url{https://github.com/XuhuiZhou/CATS}} for commonsense knowledge; and \olmpics \cite{olmpics-probe}\footnote{\url{https://github.com/alontalmor/oLMpics}} for reasoning ability.
As noted before, \nelson is the only probe package that requires additional training. All experiments in \nelson  are seeded and strictly follow the hyperparameters and training setup used in \citet{nelson-probe}. \newline
We estimate the total cost of probing: 165.54 kg CO2e carbon footprint and 345.26 kWh energy \citep{green-algo-estimator}.
\paragraph{Finetuning} For convenience and reproduciblility, we also use public package to conduct finetuning. We use \texttt{jiant}\footnote{\url{https://github.com/nyu-mll/jiant}} for ReCoRD and WSC; and Huggingface for the rest. 
Due to the large amount of training involved, we choose 14 intermediate checkpoints. In addition, it is infeasible to find optimal hyperparamerters for each individual checkpoint; to be as fair as we can, we use the same set of hyperparamters (see Table \ref{tab:hyper:finetune}) for running each finetuning task on all our checkpoints and original \roberta. This might explain the sub-optimal performance from the original \roberta in Fig.\ \ref{fig:finetune}. \newline
We estimated that fintuning emits 54.98 kg CO2e carbon footprint and consumes 121.64 kWh energy \cite{green-algo-estimator}.


\paragraph{Hardware} We run our probing and finetuning experiments on Intel Core i9-9820X CPU @ 3.30GHz and GTX 2080 Ti.




\section{Additional Probing Results}
\label{appendix:reults}
\subsection{Complete Results for Learning Patterns}
\label{appendix:results:pattern}

\begin{figure*}[thb] 
\centering
\includegraphics[scale=0.18]{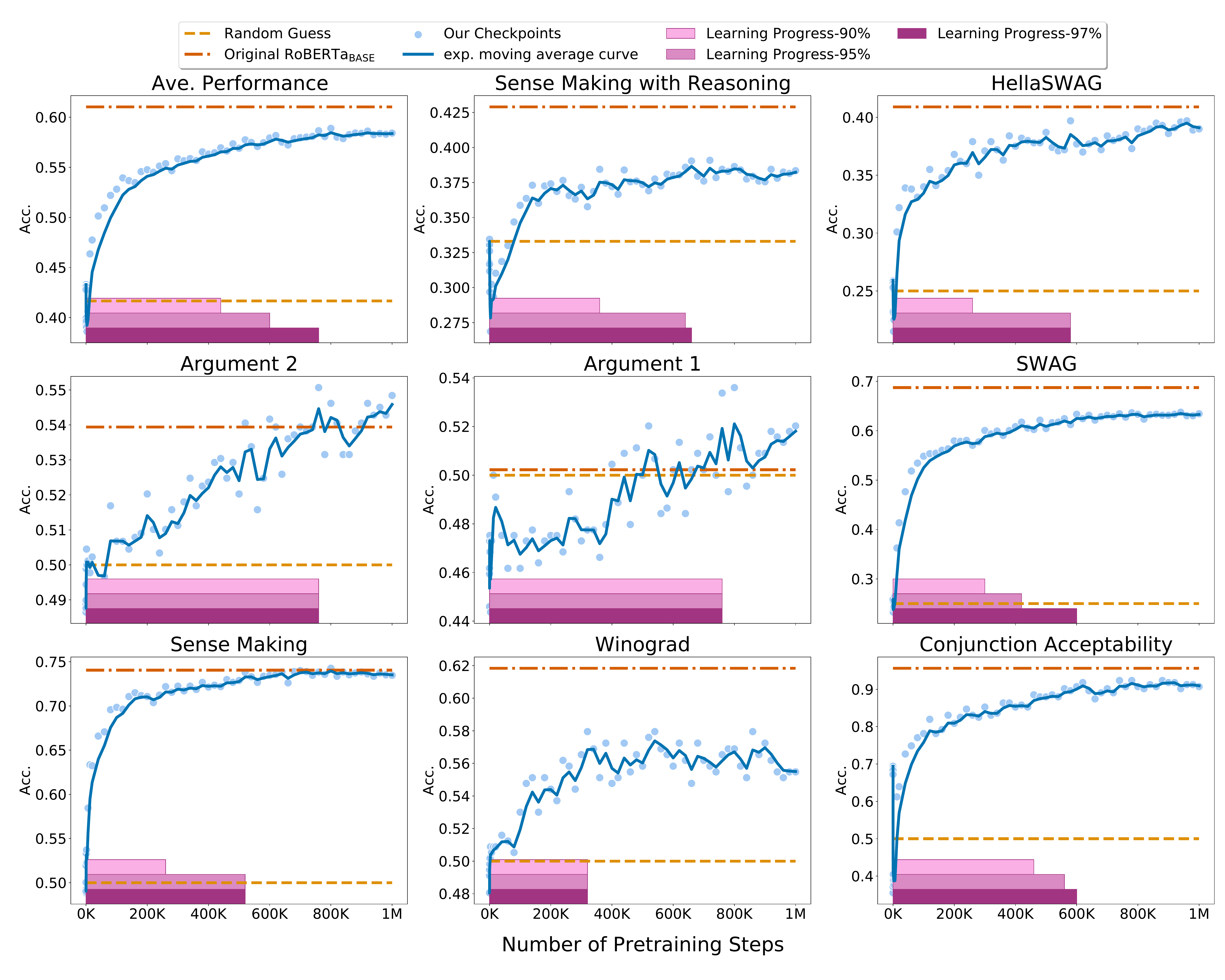}
\caption{The complete results of \textbf{\cat (Commonsense)} in Figure \ref{fig:main}, plotted in the same format.}
\label{fig:appendix:results:pattern:cat}
\end{figure*}

\begin{figure*}[thb] 
\centering
\includegraphics[scale=0.18]{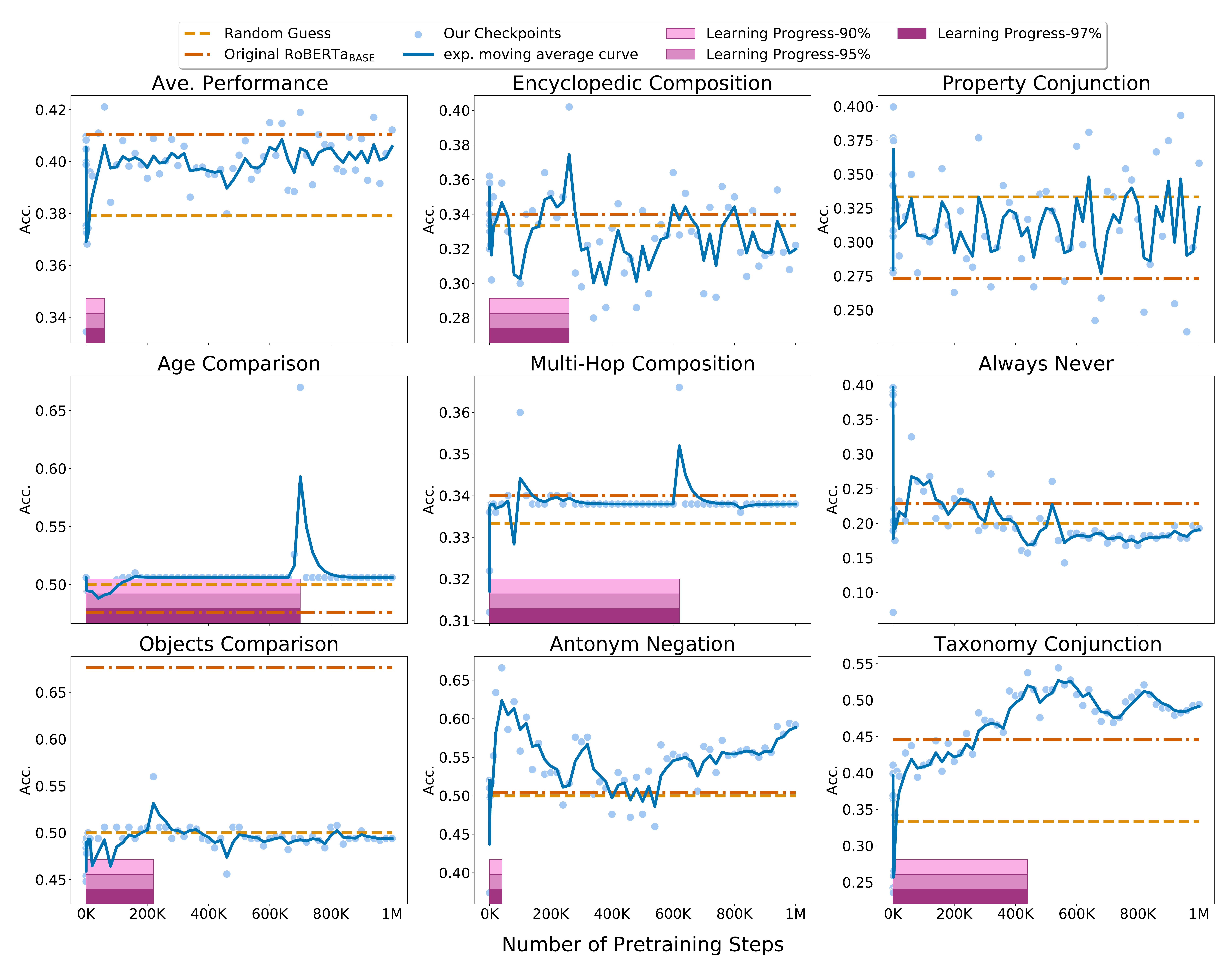}
\caption{Complete results of \textbf{\olmpics (Reasoning)} in Figure \ref{fig:main}, plotted in the same format.}
\label{fig:appendix:results:pattern:olmpics}
\end{figure*}

Here we show plots from all probing tasks we tested. Overall, the selected graphs in Fig.\ \ref{fig:main} are representative. See complete results of  \blimp in Fig.\ \ref{fig:appendix:results:pattern:blimp}, showing fast learning speed; \cat in Fig. \ref{fig:appendix:results:pattern:cat} showing slower learning speed; and \olmpics in Fig.\ \ref{fig:appendix:results:pattern:olmpics} shows a sign of ``not learning'' in many tests and ``learning'' only in a few tests.

\subsection{Correlation Plot}
\label{appendix:correlation}
\begin{figure*}[thb] 
\centering
\includegraphics[scale=0.225]{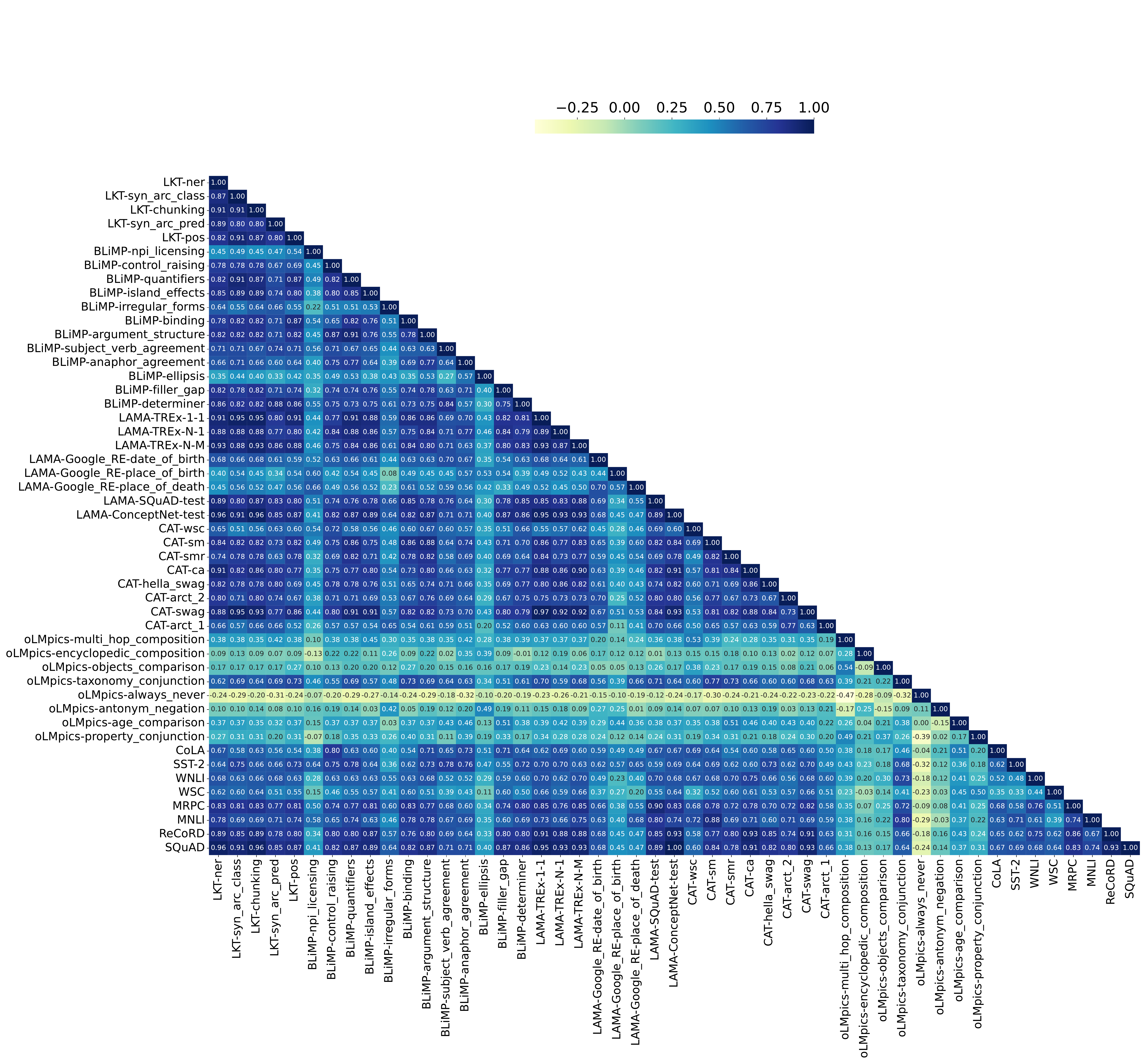}
\caption{Plot of \textbf{Kendall Tau correlation} among all experiments included. We calculate based on results from the 14 checkpoints used in Fig. \ref{fig:finetune} }
\label{fig:appendix:correlation}
\end{figure*}

\begin{figure*}[thb] 
\centering
\includegraphics[scale=0.2]{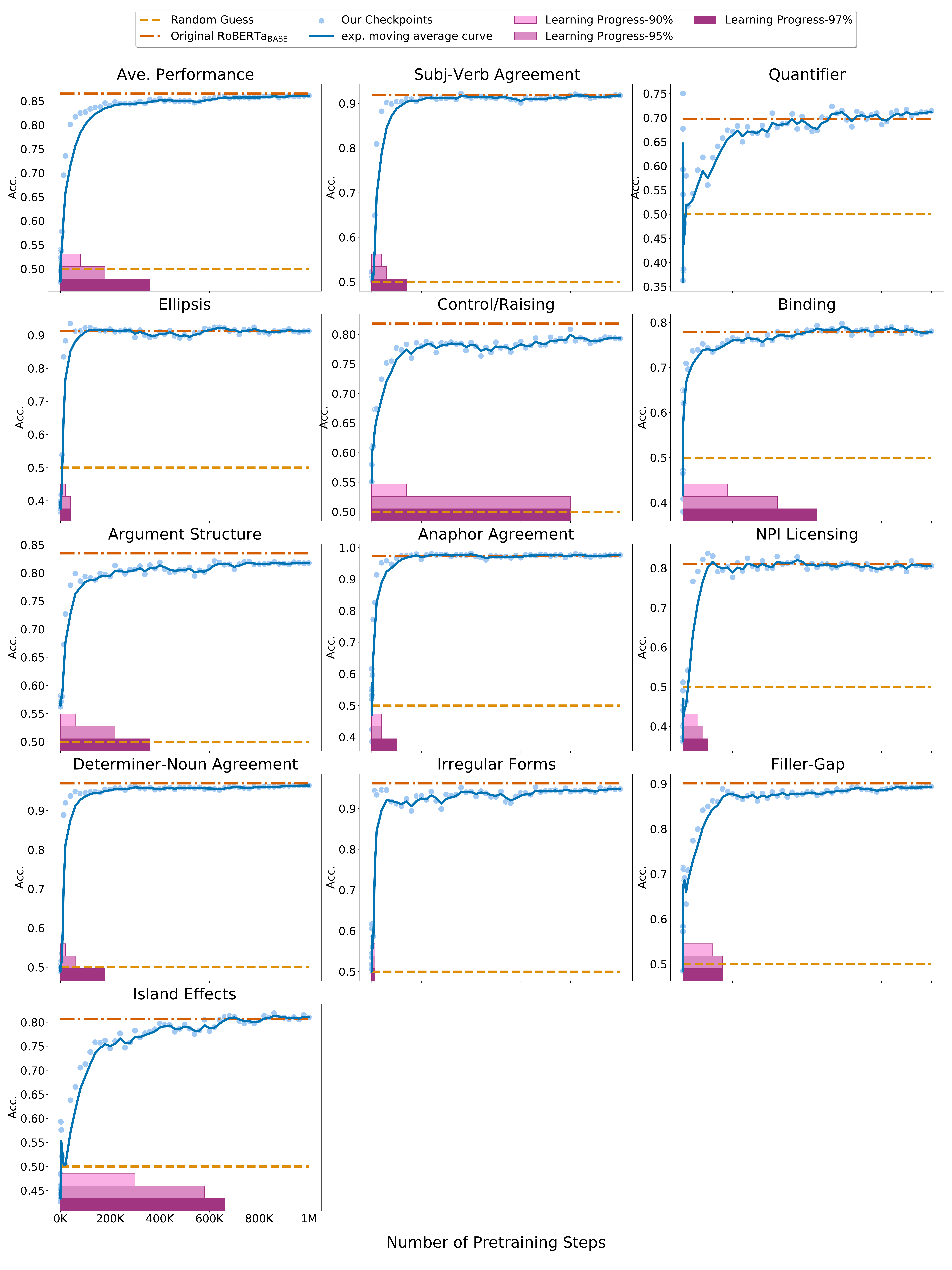}
\caption{Complete results of \textbf{\blimp (Linguistics)} in Figure \ref{fig:main}, plotted in the same format.}
\label{fig:appendix:results:pattern:blimp}
\end{figure*}

In Fig. \ref{fig:appendix:correlation}, we show a plot of correlation among all experimental results run on our replicated \roberta (\textsection \ref{sec:learning_patterns} and 
\textsection \ref{sec:finetuning}). Although successful probes usually correlate with each other, we note that low correlation with successful probes (e.g., linguistic probes) doesn't necessarily imply failure. One such example is Ellipsis in \blimp, which shows great success in Fig. \ref{fig:appendix:results:pattern:blimp}, yet has low correlation with other successful probes.

\subsection{Complete Results for Varying Pretrain Corpus Domain}
\label{appendix:results:domain}
In this section, we complement the domain results from Fig.\ \ref{fig:domains_and_sizes}. Note that the average performance shown here is calculated with all tested tasks in each probe package, whereas Fig.\ \ref{fig:domains_and_sizes} only includes the same tasks as Fig.\ \ref{fig:main} for illustration purposes. 

\begin{figure*}[thb] 
\centering
\includegraphics[scale=0.4]{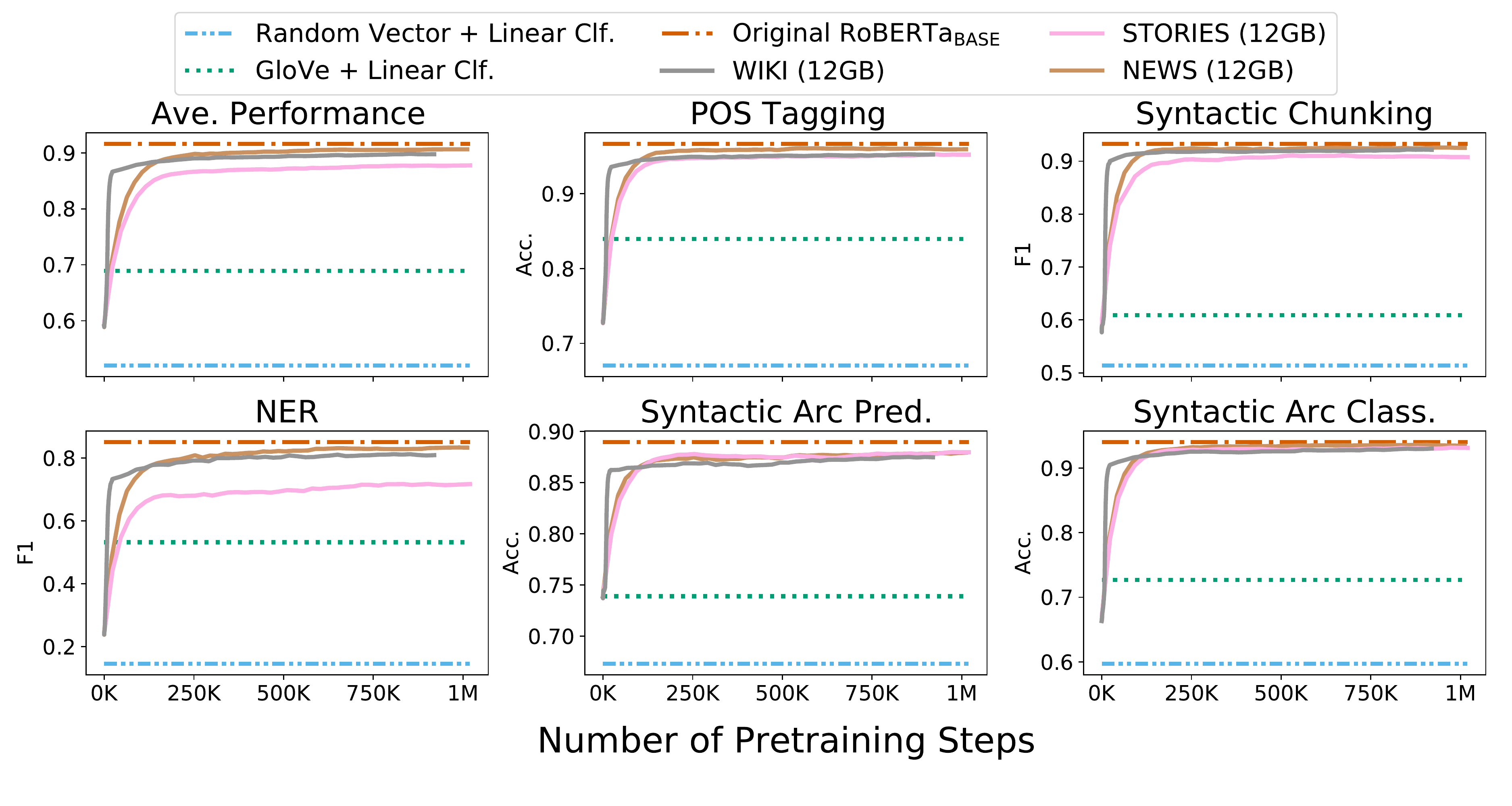}
\caption{Complete results of \textbf{\nelson(Linguistics) on different domains}. Following Fig.\ \ref{fig:main}, every line in plots are smoothed using exponential moving average with coefficient 0.5. Average performance is calculated from all included tasks in this plot.}
\label{fig:appendix:results:domain:nelson}
\end{figure*}

\begin{figure*}[thb] 
\centering
\includegraphics[scale=0.4]{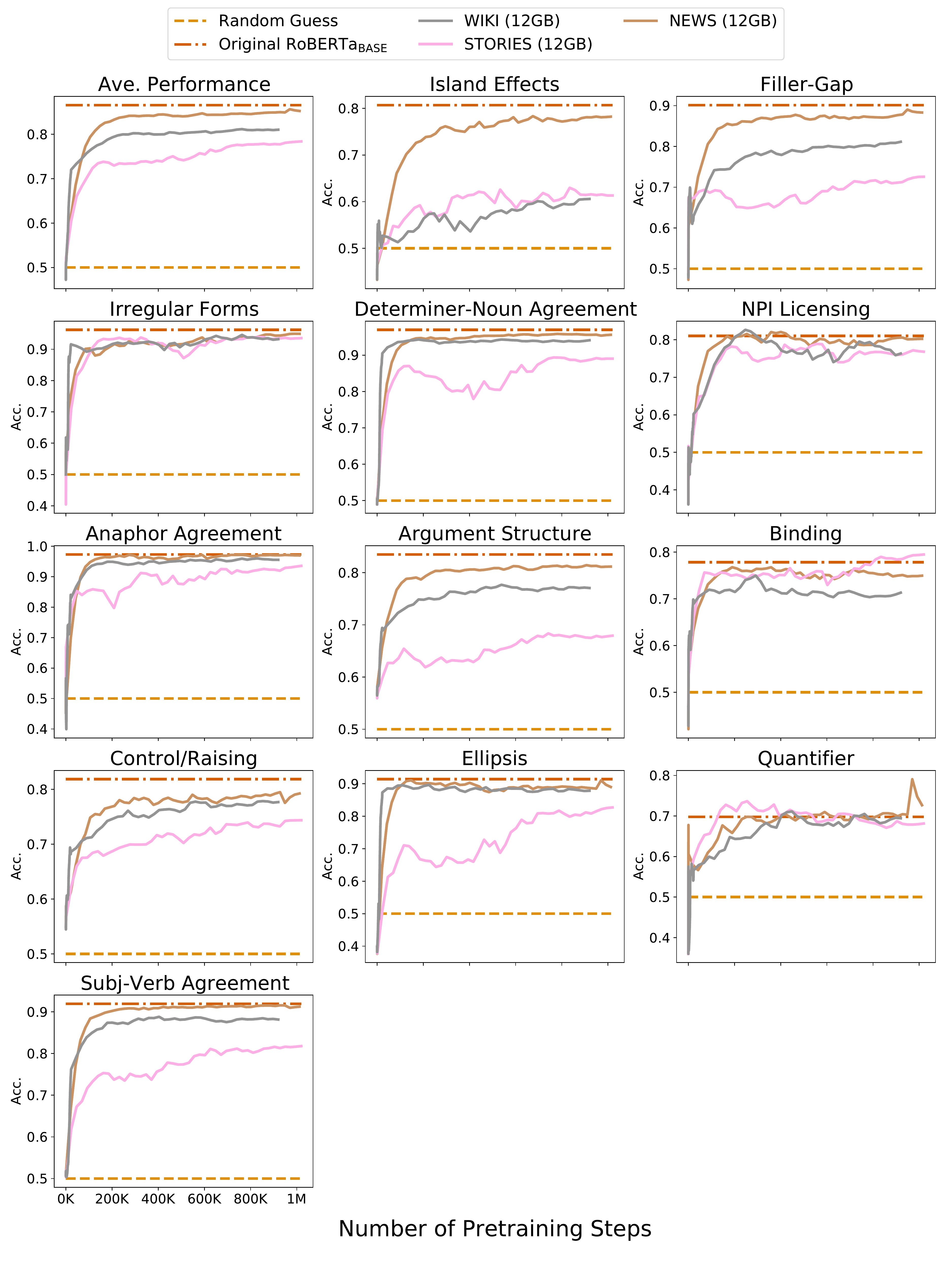}
\caption{Complete results of \textbf{\blimp(Linguistics) on different domains}. Following Fig.\ \ref{fig:main}, every line in plots are smoothed using exponential moving average with coefficient 0.5. Average performance is calculated from all included tasks in this plot.}
\label{fig:appendix:results:domain:blimp}
\end{figure*}

\begin{figure*}[thb] 
\centering
\includegraphics[scale=0.4]{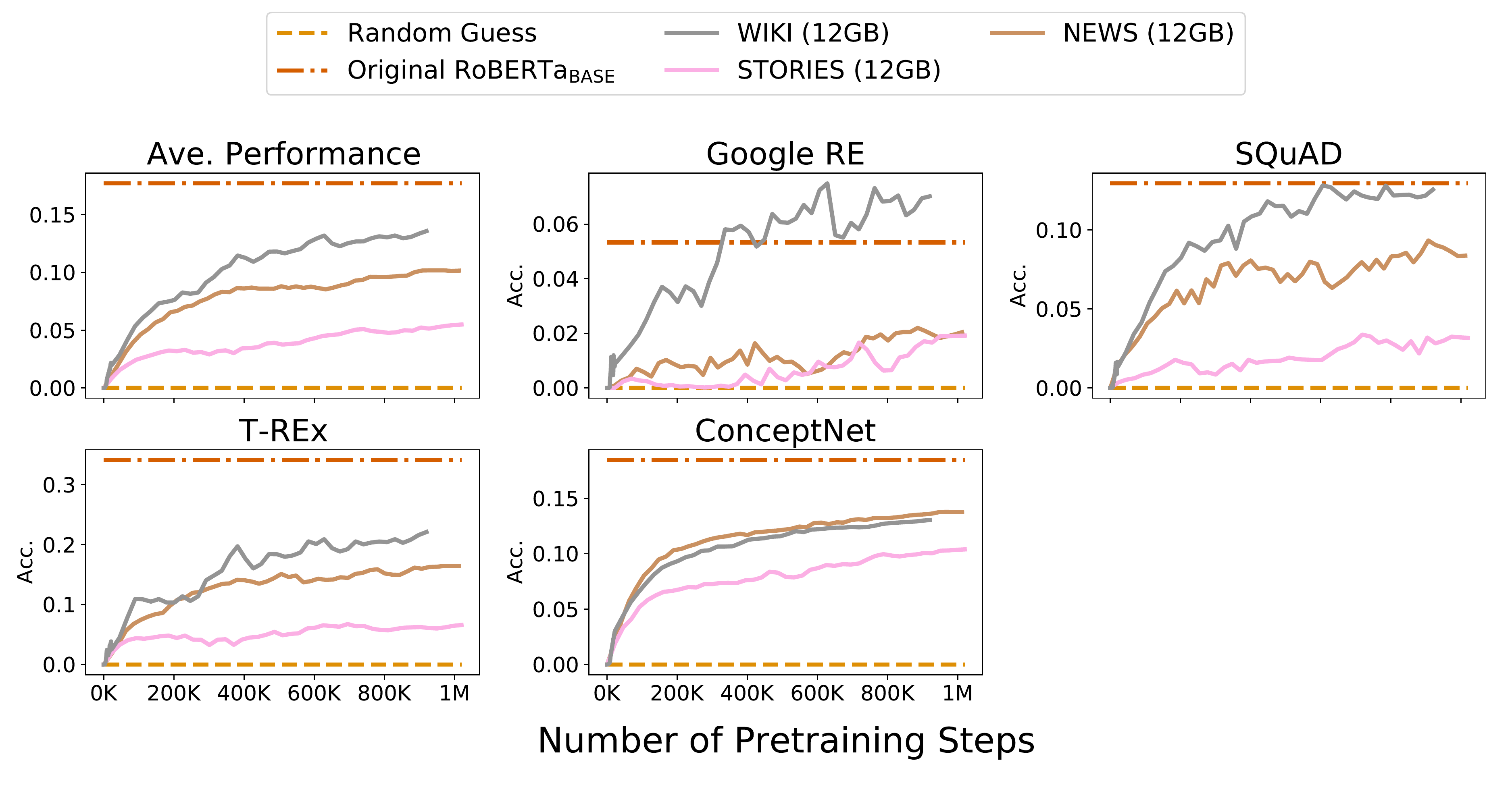}
\caption{Complete results of \textbf{\lama (Factual \& Commonsense) on different domains}. Following Fig.\ \ref{fig:main}, every line in plots are smoothed using exponential moving average with coefficient 0.5. Average performance is calculated from all included tasks in this plot.}
\label{fig:appendix:results:domain:lama}
\end{figure*}

\begin{figure*}[thb] 
\centering
\includegraphics[scale=0.4]{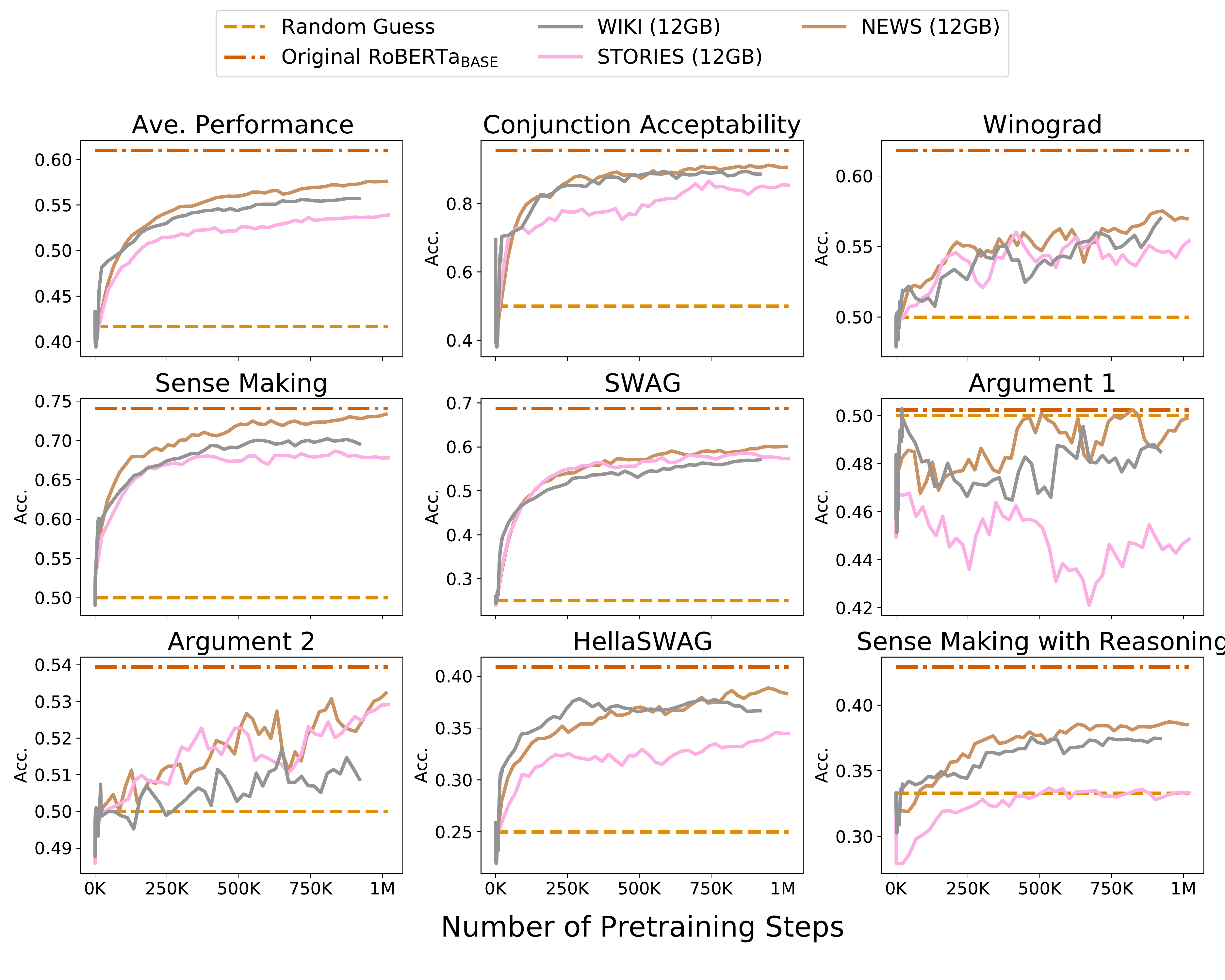}
\caption{Complete results of \textbf{\cat(Commonsense) on different domains}. Following Fig.\ \ref{fig:main}, every line in plots are smoothed using exponential moving average with coefficient 0.5. Average performance is calculated from all included tasks in this plot.}
\label{fig:appendix:results:domain:cat}
\end{figure*}

\begin{figure*}[thb] 
\centering
\includegraphics[scale=0.4]{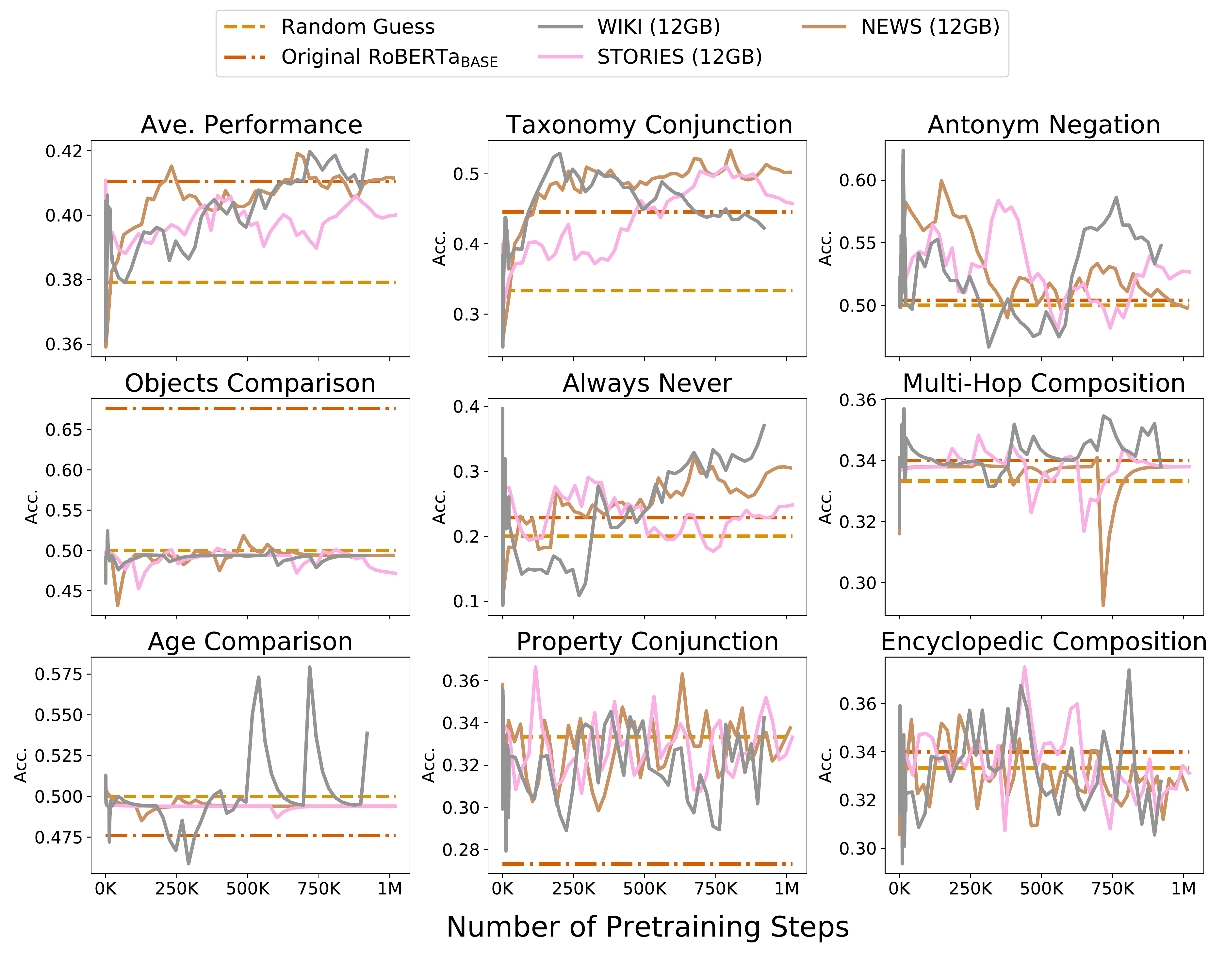}
\caption{Complete results of \textbf{\olmpics(Reasoning) on different domains}. Following Fig.\ \ref{fig:main}, every line in plots are smoothed using exponential moving average with coefficient 0.5. Average performance is calculated from all included tasks in this plot.}
\label{fig:appendix:results:domain:olmpics}
\end{figure*}

See complete results of \nelson in Fig.\ \ref{fig:appendix:results:domain:nelson} and \blimp in Fig.\ \ref{fig:appendix:results:domain:blimp} showing relatively small domain impact; \lama in Fig.\ \ref{fig:appendix:results:domain:lama} shows arguably the largest impact;  \cat in Figure \ref{fig:appendix:results:domain:cat} showing noticeable impact on some tasks; we see large variation of the impact from the domain \olmpics in Fig.\ \ref{fig:appendix:results:domain:olmpics}. Note again that \stories usually gives the slowest learning speed and worst final performance from teste probes.

\subsection{Complete Results for Varying Pretrain Corpus Size}
\label{appendix:results:size}
\begin{figure*}[thb] 
\centering
\includegraphics[scale=0.4]{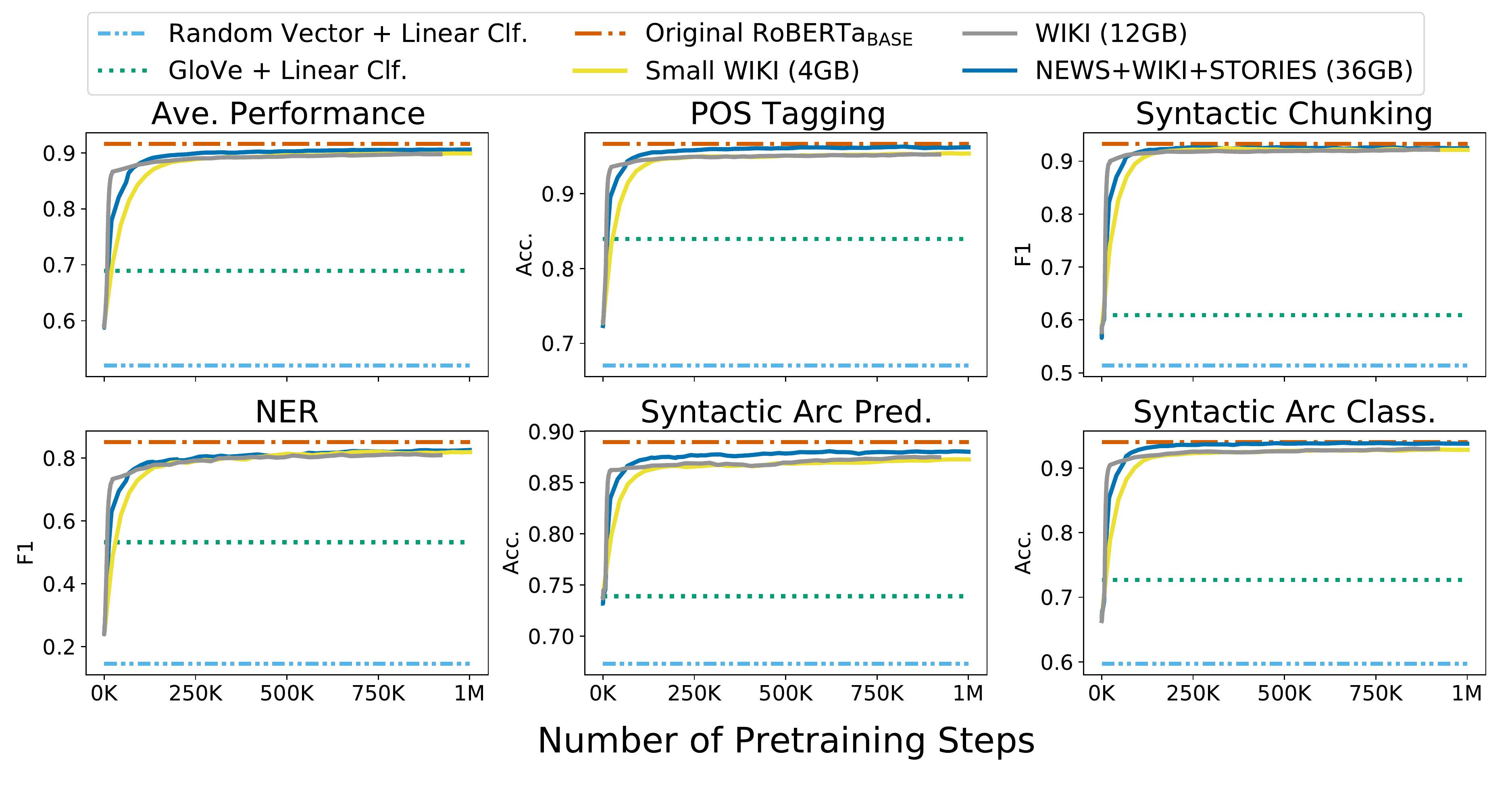}
\caption{Complete results of \textbf{\nelson (Linguistics) on different corpus sizes}. Following Fig.\ \ref{fig:main}, every line in plots are smoothed using exponential moving average with coefficient 0.5. Average performance is calculated from all included tasks in this plot.}
\label{fig:appendix:results:size:nelson}
\end{figure*}

\begin{figure*}[thb] 
\centering
\includegraphics[scale=0.4]{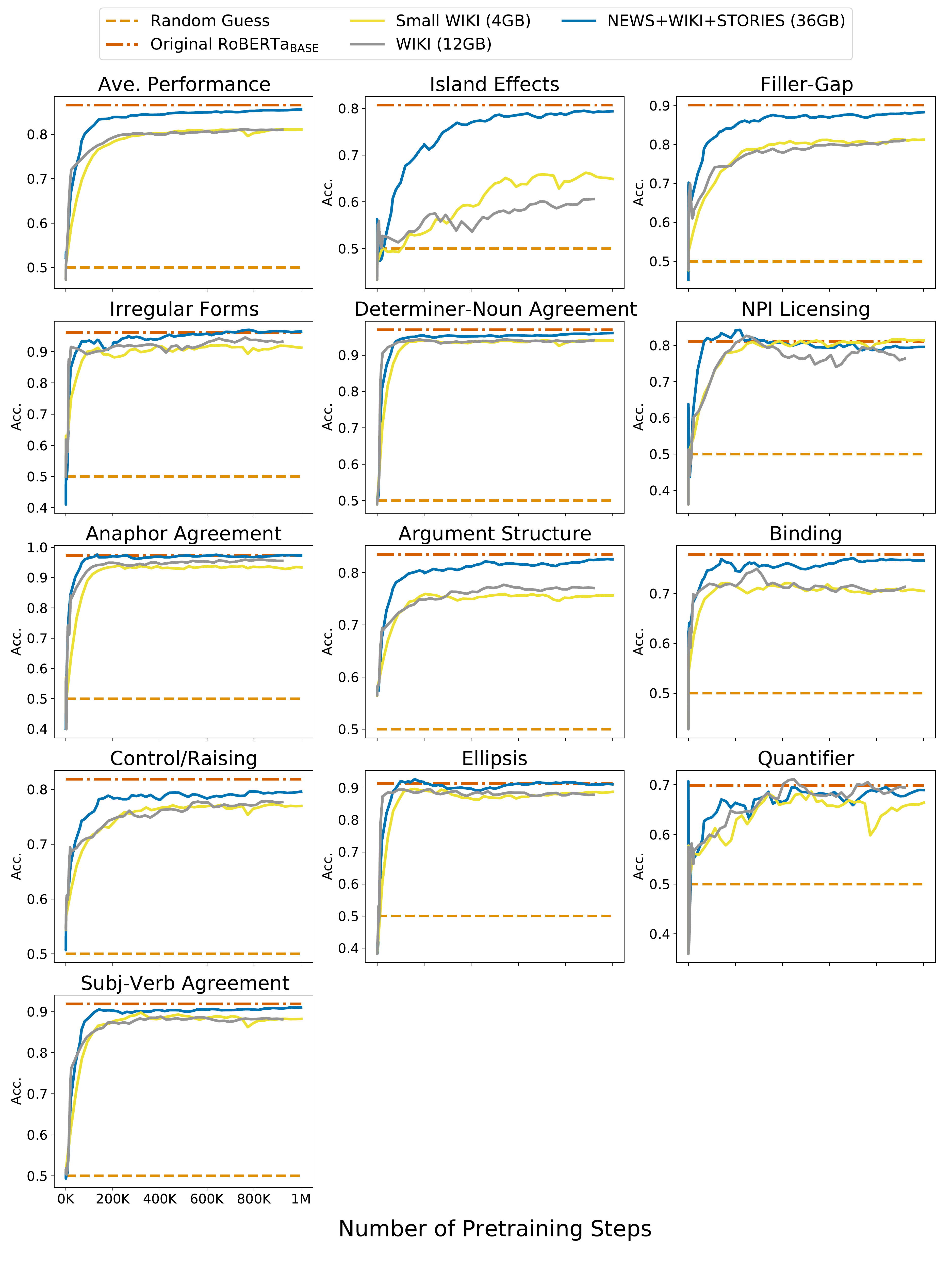}
\caption{Complete results of \textbf{\blimp (Linguistics) on different corpus sizes}. Following Fig.\ \ref{fig:main}, every line in plots are smoothed using exponential moving average with coefficient 0.5. Average performance is calculated from all included tasks in this plot.}

\label{fig:appendix:results:size:blimp}
\end{figure*}

\begin{figure*}[thb] 
\centering
\includegraphics[scale=0.4]{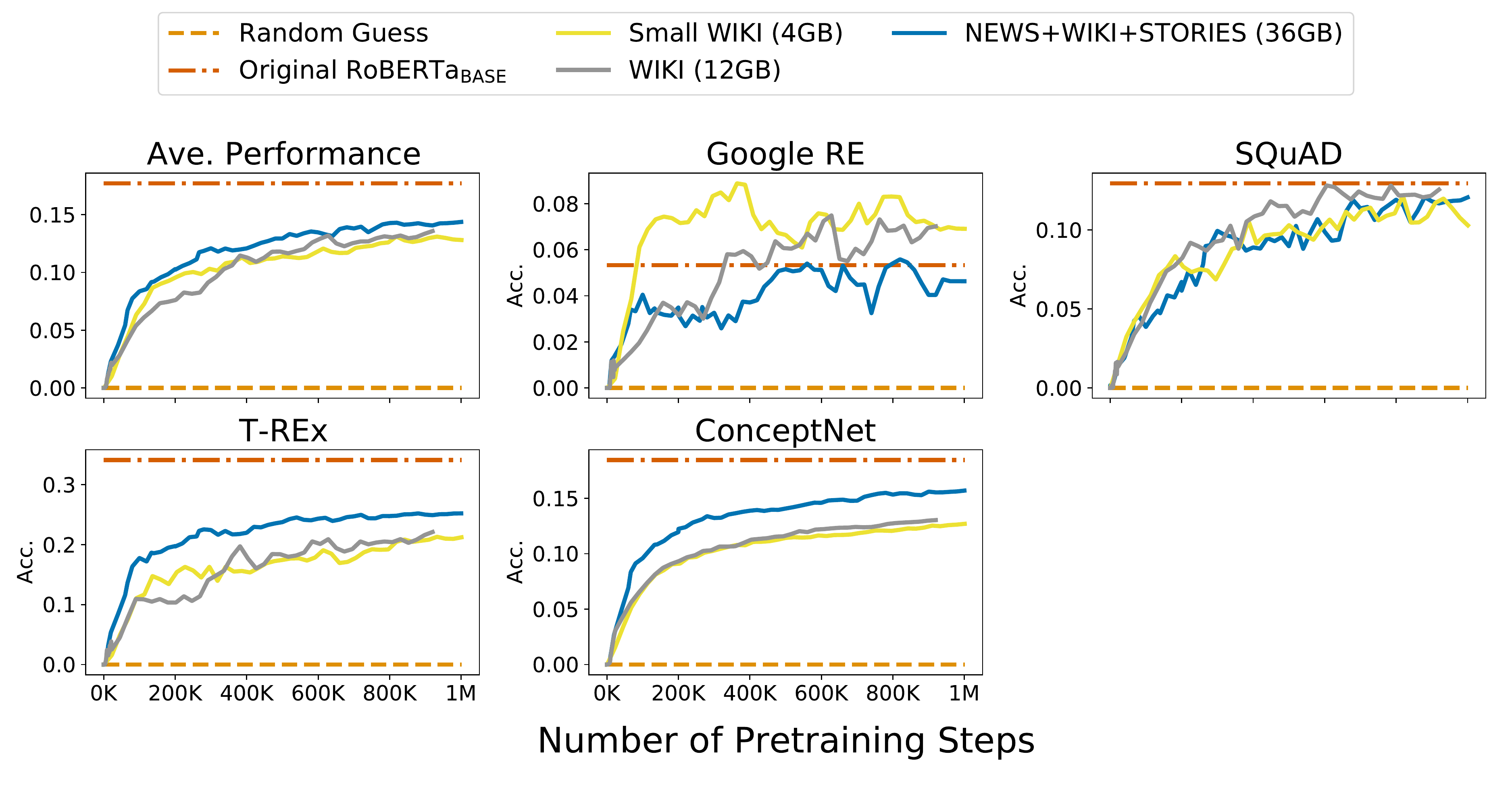}
\caption{Complete results of \textbf{\lama (Factual \& Commonsense) on different domains}. Following Fig.\ \ref{fig:main}, every line in plots are smoothed using exponential moving average with coefficient 0.5. Average performance is calculated from all included tasks in this plot.}
\label{fig:appendix:results:size:lama}
\end{figure*}

\begin{figure*}[thb] 
\centering
\includegraphics[scale=0.4]{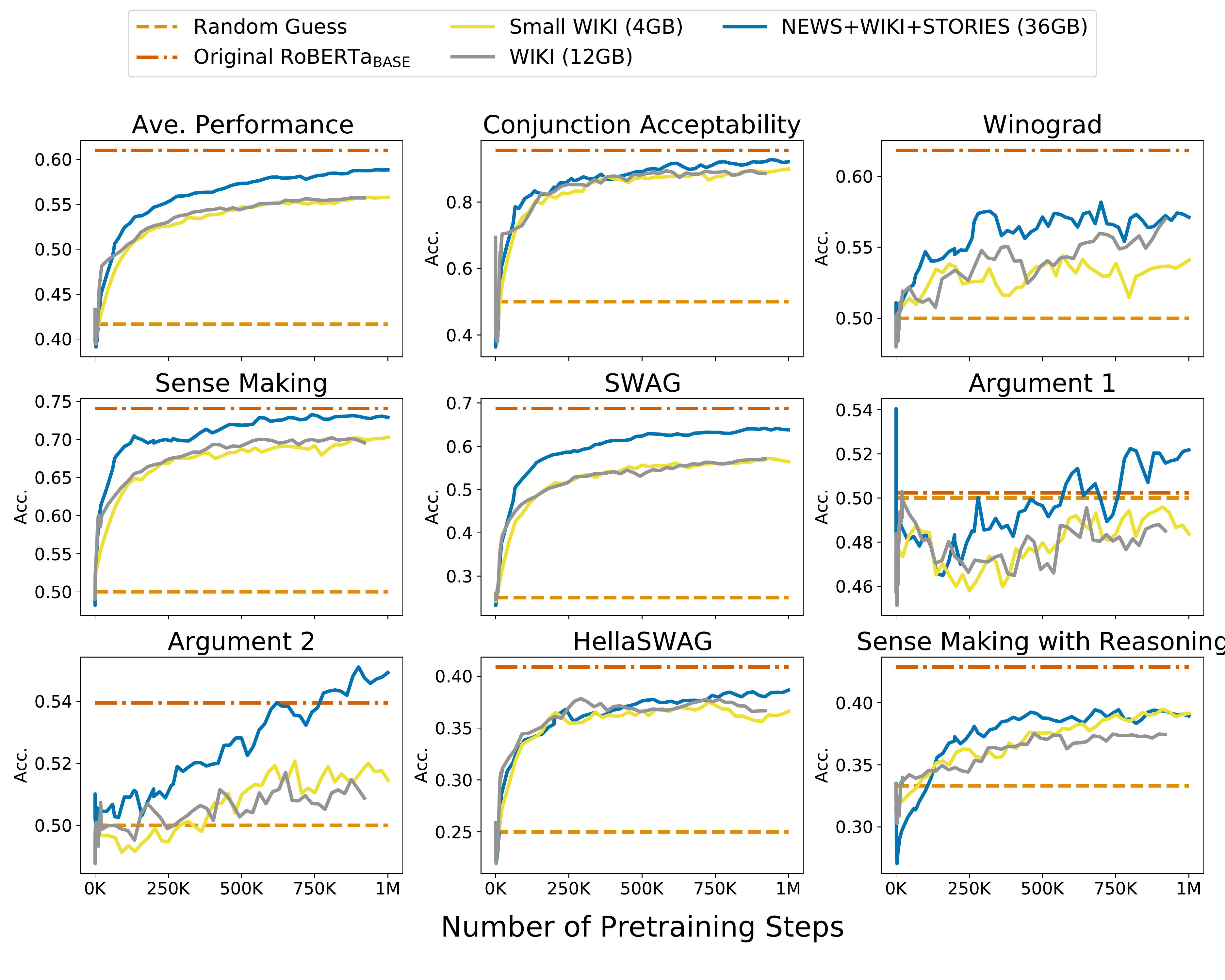}
\caption{Complete results of \textbf{\cat(Commonsense) on different corpus sizes}. Following Fig.\ \ref{fig:main}, every line in plots are smoothed using exponential moving average with coefficient 0.5. Average performance is calculated from all included tasks in this plot.}
\label{fig:appendix:results:size:cat}
\end{figure*}

\begin{figure*}[thb] 
\centering
\includegraphics[scale=0.4]{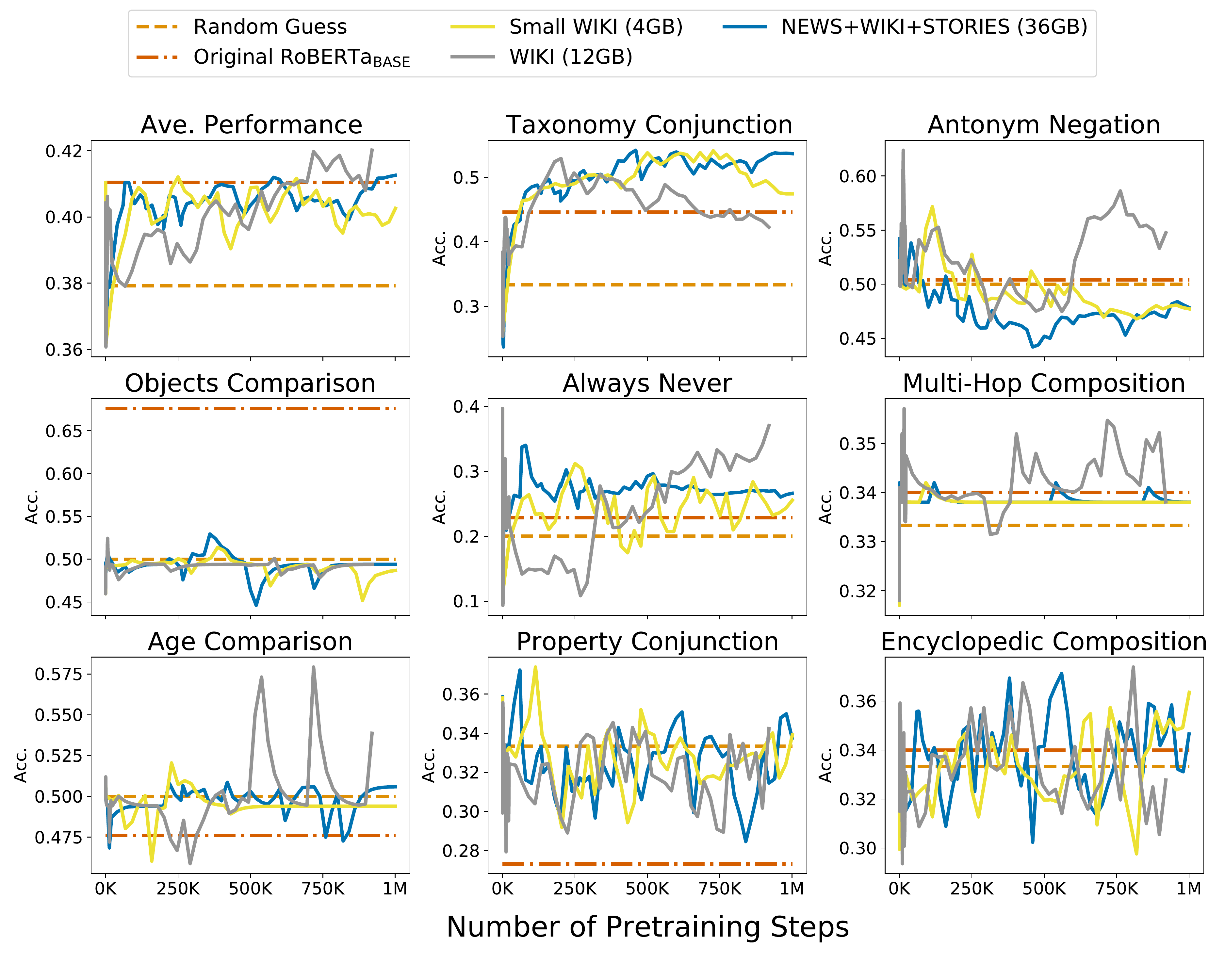}
\caption{Complete results of \textbf{\olmpics(Reasoning) on different corpus sizes}. Following Fig.\ \ref{fig:main}, every line in plots are smoothed using exponential moving average with coefficient 0.5. Average performance is calculated from all included tasks in this plot.}
\label{fig:appendix:results:size:olmpics}
\end{figure*}

In this section, we complement the corpus size results from Fig.\ \ref{fig:domains_and_sizes}. Note that the average performance shown here is calculated with all included tasks in each probe package, whereas Fig.\ \ref{fig:domains_and_sizes}'s use the same tasks as Fig.\ \ref{fig:main} for illustration purpose. 
See complete results of \nelson in Fig.\ \ref{fig:appendix:results:size:nelson}; \blimp in Fig.\ \ref{fig:appendix:results:size:blimp}; \lama in Fig.\ \ref{fig:appendix:results:size:lama}; \cat in Fig.\ \ref{fig:appendix:results:size:cat}; \olmpics in Fig.\ \ref{fig:appendix:results:size:olmpics}. 
To reiterate our conclusion from \textsection \ref{sec:domain_and_size}, the biggest corpus (with more diverse inclusion of domains) generally learns faster and results in an ultimately better \roberta in all tested knowledge; in contrast, we don't observe this when we simply change from downsampled English \wiki (4 GB) to original English \wiki (12 GB).

\input{tables/tab_hyper_pretrain}
\input{tables/tab_num_checkpoints}
\input{tables/tab_hyper_finetune}









%% file: tables/tab_hyper_pretrain.tex
\begin{table*}[tbh] 
\centering
    \begin{tabular}{cc}
        \toprule
        \textbf{Name} & \textbf{Values}  \\
        \toprule
            Architecture & RoBERTa-base  \\
            Masking & Static \\
            Update steps & 1M   \\
            Batch size & 256 \\
            Max length & 512 \\
            Warmup steps & 10K \\
            Peak Learning rate & 0.0005 \\
            Learning rate scheduler & Polynomial Decay \\
            Dropout rate & 0.1 \\
            Attention dropout rate & 0.1\\
            Weight Decay rate & 0.01 \\
            Optimizer & Adam$(\beta$s $= (0.9, 0.98)$, $\epsilon=1e-6)$ \\
        \bottomrule
    \end{tabular}
\caption{Configurations for pretraining \roberta.}
\label{tab:hyper:pretrain}
\end{table*}

%% file: tables/tab_num_checkpoints.tex
\begin{table*}[tbh]
\small
\centering
\setlength\tabcolsep{8pt}   
\def\arraystretch{1.8}      
\begin{tabular}{c|c}
\toprule
\textbf{Domain}          & \textbf{No. of Checkpoints} \\
\cline{1-2}
NEWS+WIKI+STORIES (36GB) & 65                          \\
NEWS (12GB)              & 56                          \\
STORIES (12GB)           & 52                          \\
WIKI (12GB)              & 71                          \\
Small WIKI (4GB)         & 52                          \\
Reproduced               & 62                         \\
\bottomrule
\end{tabular}
\caption{Numbers of saved checkpoints during pretraining \roberta on different domains. ``Reproduced'' denotes our replication of model in \cite{liu_roberta:_2019}, detailed in \textsection \ref{sec:method:pretraining}. The rest corresponds to corpora experimented on basis of domains and corpus sizes, detailed in \textsection \ref{sec:domain_and_size}.}
\label{tab:num-checkpoints}
\end{table*}

%% file: tables/tab_hyper_finetune.tex
\begin{table*}[tbh]
\begin{tabular}{ccccccccc}
\toprule
                         & CoLA & SST-2 & MRPC & WNLI & MNLI & SQuAD & WSC  & ReCoRD \\ \toprule
\textbf{Batch size}      & 32   & 32    & 32   & 32   & 32   & 12    & 32   & 32     \\
\textbf{Epoch}           & 3    & 3     & 3    & 3    & 3    & 2     & 10   & 2      \\
\textbf{Learning rate}   & 2e-5 & 2e-5  & 2e-5 & 2e-5 & 2e-5 & 3e-5  & 1e-5 & 1e-5   \\
\bottomrule
\end{tabular}
\caption{Important hyperparamters for finetuning experiments in \textsection \ref{sec:finetuning}.  Experiments are seeded for reproducibility.}
\label{tab:hyper:finetune}
\end{table*}

%% file: main.bbl
\begin{thebibliography}{40}
\expandafter\ifx\csname natexlab\endcsname\relax\def\natexlab#1{#1}\fi

\bibitem[{Belinkov et~al.(2020)Belinkov, Gehrmann, and
  Pavlick}]{belinkov-etal-2020-interpretability-tutorial}
Yonatan Belinkov, Sebastian Gehrmann, and Ellie Pavlick. 2020.
\newblock \href {https://www.aclweb.org/anthology/2020.acl-tutorials.1}
  {Interpretability and analysis in neural {NLP}}.
\newblock In \emph{Proc.\ of ACL: Tutorial Abstracts}.

\bibitem[{Bhagavatula et~al.(2020)Bhagavatula, Bras, Malaviya, Sakaguchi,
  Holtzman, Rashkin, Downey, Yih, and Choi}]{abdutive-commonsense-reasoning}
Chandra Bhagavatula, Ronan~Le Bras, Chaitanya Malaviya, Keisuke Sakaguchi, Ari
  Holtzman, Hannah Rashkin, Doug Downey, Wen{-}tau Yih, and Yejin Choi. 2020.
\newblock \href {https://openreview.net/forum?id=Byg1v1HKDB} {Abductive
  commonsense reasoning}.
\newblock In \emph{Proc.\ of ICLR}.

\bibitem[{Brown et~al.(2020)Brown, Mann, Ryder, Subbiah, Kaplan, Dhariwal,
  Neelakantan, Shyam, Sastry, Askell, Agarwal, Herbert{-}Voss, Krueger,
  Henighan, Child, Ramesh, Ziegler, Wu, Winter, Hesse, Chen, Sigler, Litwin,
  Gray, Chess, Clark, Berner, McCandlish, Radford, Sutskever, and
  Amodei}]{gpt3}
Tom~B. Brown, Benjamin Mann, Nick Ryder, Melanie Subbiah, Jared Kaplan,
  Prafulla Dhariwal, Arvind Neelakantan, Pranav Shyam, Girish Sastry, Amanda
  Askell, Sandhini Agarwal, Ariel Herbert{-}Voss, Gretchen Krueger, Tom
  Henighan, Rewon Child, Aditya Ramesh, Daniel~M. Ziegler, Jeffrey Wu, Clemens
  Winter, Christopher Hesse, Mark Chen, Eric Sigler, Mateusz Litwin, Scott
  Gray, Benjamin Chess, Jack Clark, Christopher Berner, Sam McCandlish, Alec
  Radford, Ilya Sutskever, and Dario Amodei. 2020.
\newblock \href {https://arxiv.org/abs/2005.14165} {Language models are
  few-shot learners}.

\bibitem[{Chiang et~al.(2020)Chiang, Huang, and Lee}]{serious-idea-crash}
Cheng-Han Chiang, Sung-Feng Huang, and Hung-yi Lee. 2020.
\newblock \href {https://doi.org/10.18653/v1/2020.emnlp-main.553} {{P}retrained
  language model embryology: {T}he birth of {ALBERT}}.
\newblock In \emph{Proceedings of the 2020 Conference on Empirical Methods in
  Natural Language Processing (EMNLP)}, pages 6813--6828, Online. Association
  for Computational Linguistics.

\bibitem[{Clark et~al.(2019)Clark, Khandelwal, Levy, and
  Manning}]{bert-look-at}
Kevin Clark, Urvashi Khandelwal, Omer Levy, and Christopher~D Manning. 2019.
\newblock \href {https://arxiv.org/abs/1906.04341} {What does {BERT} look at?
  an analysis of {BERT}’s attention}.
\newblock In \emph{Proc.\ of BlackboxNLP}.

\bibitem[{Devlin et~al.(2019)Devlin, Chang, Lee, and
  Toutanova}]{devlin_bert:_2018}
Jacob Devlin, Ming-Wei Chang, Kenton Lee, and Kristina Toutanova. 2019.
\newblock \href {http://arxiv.org/abs/1810.04805} {{BERT}: {Pre}-training of
  {Deep} {Bidirectional} {Transformers} for {Language} {Understanding}}.
\newblock In \emph{Proc.\ of NAACL}.

\bibitem[{Dolan and Brockett(2005)}]{MRPC}
William~B. Dolan and Chris Brockett. 2005.
\newblock \href {https://www.aclweb.org/anthology/I05-5002/} {Automatically
  constructing a corpus of sentential paraphrases}.
\newblock In \emph{Proc.\ of IWP}. Asian Federation of Natural Language
  Processing.

\bibitem[{Gokaslan and Cohen(2019)}]{openwebtext-dataset}
Aaron Gokaslan and Vanya Cohen. 2019.
\newblock {OpenWebText} corpus.
\newblock \url{http://Skylion007.github.io/OpenWebTextCorpus}.

\bibitem[{Gururangan et~al.(2020)Gururangan, Marasovi{\'c}, Swayamdipta, Lo,
  Beltagy, Downey, and Smith}]{gururangan-etal-2020-dont}
Suchin Gururangan, Ana Marasovi{\'c}, Swabha Swayamdipta, Kyle Lo, Iz~Beltagy,
  Doug Downey, and Noah~A. Smith. 2020.
\newblock \href {https://www.aclweb.org/anthology/2020.acl-main.740} {Don{'}t
  stop pretraining: Adapt language models to domains and tasks}.
\newblock In \emph{Proc. of ACL}.

\bibitem[{Hao et~al.(2019)Hao, Dong, Wei, and Xu}]{visualizing-bert}
Yaru Hao, Li~Dong, Furu Wei, and Ke~Xu. 2019.
\newblock \href {https://arxiv.org/abs/1908.05620} {Visualizing and
  understanding the effectiveness of {BERT}}.
\newblock In \emph{Proc. of EMNLP}.

\bibitem[{Hewitt and Liang(2019)}]{hewitt-probe-control}
John Hewitt and Percy Liang. 2019.
\newblock \href {https://doi.org/10.18653/v1/D19-1275} {Designing and
  interpreting probes with control tasks}.
\newblock In \emph{Proc.\ of EMNLP}. Association for Computational Linguistics.

\bibitem[{Kovaleva et~al.(2019)Kovaleva, Romanov, Rogers, and
  Rumshisky}]{revealing-secrets-bert}
Olga Kovaleva, Alexey Romanov, Anna Rogers, and Anna Rumshisky. 2019.
\newblock Revealing the dark secrets of bert.
\newblock In \emph{Proc. of EMNLP}.

\bibitem[{Lannelongue et~al.(2020)Lannelongue, Grealey, and
  Inouye}]{green-algo-estimator}
Lo{\"{\i}}c Lannelongue, Jason Grealey, and Michael Inouye. 2020.
\newblock \href {http://arxiv.org/abs/2007.07610} {Green algorithms:
  Quantifying the carbon emissions of computation}.
\newblock \emph{CoRR}, abs/2007.07610.

\bibitem[{Levesque et~al.(2012)Levesque, Davis, and
  Morgenstern}]{winograd-schema-challenge}
Hector~J. Levesque, Ernest Davis, and Leora Morgenstern. 2012.
\newblock \href {http://www.aaai.org/ocs/index.php/KR/KR12/paper/view/4492}
  {The {Winograd} schema challenge}.
\newblock In \emph{Proc.\ of KR\&R}.

\bibitem[{Liu et~al.(2019{\natexlab{a}})Liu, Gardner, Belinkov, Peters, and
  Smith}]{nelson-probe}
Nelson~F. Liu, Matt Gardner, Yonatan Belinkov, Matthew~E. Peters, and Noah~A.
  Smith. 2019{\natexlab{a}}.
\newblock \href {https://doi.org/10.18653/v1/n19-1112} {Linguistic knowledge
  and transferability of contextual representations}.
\newblock In \emph{Proc.\ of NAACL}.

\bibitem[{Liu et~al.(2019{\natexlab{b}})Liu, Ott, Goyal, Du, Joshi, Chen, Levy,
  Lewis, Zettlemoyer, and Stoyanov}]{liu_roberta:_2019}
Yinhan Liu, Myle Ott, Naman Goyal, Jingfei Du, Mandar Joshi, Danqi Chen, Omer
  Levy, Mike Lewis, Luke Zettlemoyer, and Veselin Stoyanov. 2019{\natexlab{b}}.
\newblock \href {http://arxiv.org/abs/1907.11692} {{RoBERTa}: {A} {Robustly}
  {Optimized} {BERT} {Pretraining} {Approach}}.

\bibitem[{McClelland and Rumelhart(1986)}]{rumelhart}
James~L. McClelland and David~E. Rumelhart. 1986.
\newblock \href
  {https://mitpress.mit.edu/books/parallel-distributed-processing-volume-2}
  {\emph{Parallel Distributed Processing}}, volume~2.

\bibitem[{Mostafazadeh et~al.(2016)Mostafazadeh, Chambers, He, Parikh, Batra,
  Vanderwende, Kohli, and Allen}]{ROCStories}
Nasrin Mostafazadeh, Nathanael Chambers, Xiaodong He, Devi Parikh, Dhruv Batra,
  Lucy Vanderwende, Pushmeet Kohli, and James~F. Allen. 2016.
\newblock \href {https://doi.org/10.18653/v1/n16-1098} {A corpus and cloze
  evaluation for deeper understanding of commonsense stories}.
\newblock In \emph{Proc.\ of NAACL}.

\bibitem[{Nagel(2016)}]{ccnews-dataset}
Sebastian Nagel. 2016.
\newblock {CC-News}.
\newblock
  \url{http://web.archive.org/save/http://commoncrawl.org/2016/10/news-dataset-availabl}.

\bibitem[{Pennington et~al.(2014)Pennington, Socher, and
  Manning}]{pennington-etal-2014-glove}
Jeffrey Pennington, Richard Socher, and Christopher Manning. 2014.
\newblock \href {https://doi.org/10.3115/v1/D14-1162} {{G}lo{V}e: Global
  vectors for word representation}.
\newblock In \emph{Proc.\ of EMNLP}.

\bibitem[{Peters et~al.(2018)Peters, Neumann, Iyyer, Gardner, Clark, Lee, and
  Zettlemoyer}]{elmo}
Matthew~E. Peters, Mark Neumann, Mohit Iyyer, Matt Gardner, Christopher Clark,
  Kenton Lee, and Luke Zettlemoyer. 2018.
\newblock \href {https://doi.org/10.18653/v1/n18-1202} {Deep contextualized
  word representations}.
\newblock In \emph{Proc.\ of NAACL}.

\bibitem[{Petroni et~al.(2019)Petroni, Rockt{\"{a}}schel, Riedel, Lewis,
  Bakhtin, Wu, and Miller}]{lama-probe}
Fabio Petroni, Tim Rockt{\"{a}}schel, Sebastian Riedel, Patrick S.~H. Lewis,
  Anton Bakhtin, Yuxiang Wu, and Alexander~H. Miller. 2019.
\newblock \href {https://doi.org/10.18653/v1/D19-1250} {Language models as
  knowledge bases?}
\newblock In \emph{Proc.\ of EMNLP}.

\bibitem[{Pimentel et~al.(2020)Pimentel, Valvoda, Maudslay, Zmigrod, Williams,
  and Cotterell}]{pimentel-probe-information}
Tiago Pimentel, Josef Valvoda, Rowan~Hall Maudslay, Ran Zmigrod, Adina
  Williams, and Ryan Cotterell. 2020.
\newblock \href {https://arxiv.org/abs/2004.03061} {Information-theoretic
  probing for linguistic structure}.
\newblock In \emph{Proc.\ of ACL}.

\bibitem[{Qin et~al.(2020)Qin, Shwartz, West, Bhagavatula, Hwang, Bras,
  Bosselut, and Choi}]{back-to-the-future}
Lianhui Qin, Vered Shwartz, Peter West, Chandra Bhagavatula, Jena~D. Hwang,
  Ronan~Le Bras, Antoine Bosselut, and Yejin Choi. 2020.
\newblock \href {https://www.aclweb.org/anthology/2020.emnlp-main.58/} {Back to
  the future: Unsupervised backprop-based decoding for counterfactual and
  abductive commonsense reasoning}.
\newblock In \emph{Proc.\ of EMNLP}.

\bibitem[{Radford et~al.(2018)Radford, Wu, Child, Luan, Amodei, and
  Sutskever}]{gpt2}
Alec Radford, Jeffrey Wu, Rewon Child, David Luan, Dario Amodei, and Ilya
  Sutskever. 2018.
\newblock \href
  {https://d4mucfpksywv.cloudfront.net/better-language-models/language-models.pdf}
  {Language models are unsupervised multitask learners}.

\bibitem[{Raffel et~al.(2020)Raffel, Shazeer, Roberts, Lee, Narang, Matena,
  Zhou, Li, and Liu}]{raffel-etal-2020-t5}
Colin Raffel, Noam Shazeer, Adam Roberts, Katherine Lee, Sharan Narang, Michael
  Matena, Yanqi Zhou, W.~Li, and Peter~J. Liu. 2020.
\newblock \href {https://arxiv.org/abs/1910.10683} {Exploring the limits of
  transfer learning with a unified text-to-text transformer}.
\newblock \emph{JMLR}.

\bibitem[{Rajpurkar et~al.(2016)Rajpurkar, Zhang, Lopyrev, and
  Liang}]{squad1.1-dataset}
Pranav Rajpurkar, Jian Zhang, Konstantin Lopyrev, and Percy Liang. 2016.
\newblock \href {https://doi.org/10.18653/v1/d16-1264} {Squad: 100, 000+
  questions for machine comprehension of text}.
\newblock In \emph{Proc.\ of EMNLP}.

\bibitem[{Salazar et~al.(2020)Salazar, Liang, Nguyen, and
  Kirchhoff}]{salazar-etal-2020-masked}
Julian Salazar, Davis Liang, Toan~Q. Nguyen, and Katrin Kirchhoff. 2020.
\newblock \href {https://doi.org/10.18653/v1/2020.acl-main.240} {Masked
  language model scoring}.
\newblock In \emph{Proc.\ of ACL}.

\bibitem[{Saphra and Lopez(2019)}]{dynamics-by-svcca}
Naomi Saphra and Adam Lopez. 2019.
\newblock \href {https://doi.org/10.18653/v1/n19-1329} {Understanding learning
  dynamics of language models with {SVCCA}}.
\newblock In \emph{Proc.\ of NAACL}.

\bibitem[{Socher et~al.(2013)Socher, Perelygin, Wu, Chuang, Manning, Ng, and
  Potts}]{sst2-dataset}
Richard Socher, Alex Perelygin, Jean Wu, Jason Chuang, Christopher~D. Manning,
  Andrew~Y. Ng, and Christopher Potts. 2013.
\newblock \href {https://www.aclweb.org/anthology/D13-1170/} {Recursive deep
  models for semantic compositionality over a sentiment treebank}.
\newblock In \emph{Proc.\ of EMNLP}.

\bibitem[{Talmor et~al.(2019)Talmor, Elazar, Goldberg, and
  Berant}]{olmpics-probe}
Alon Talmor, Yanai Elazar, Yoav Goldberg, and Jonathan Berant. 2019.
\newblock \href {http://arxiv.org/abs/1912.13283} {{oLMpics} - on what language
  model pre-training captures}.
\newblock \emph{TACL}.

\bibitem[{Trinh and Le(2018)}]{stories-dataset}
Trieu~H. Trinh and Quoc~V. Le. 2018.
\newblock \href {http://arxiv.org/abs/1806.02847} {A simple method for
  commonsense reasoning}.

\bibitem[{Warstadt et~al.(2020)Warstadt, Parrish, Liu, Mohananey, Peng, Wang,
  and Bowman}]{blimp-probe}
Alex Warstadt, Alicia Parrish, Haokun Liu, Anhad Mohananey, Wei Peng,
  Sheng{-}Fu Wang, and Samuel~R. Bowman. 2020.
\newblock \href {https://transacl.org/ojs/index.php/tacl/article/view/2013}
  {{BLiMP}: The benchmark of linguistic minimal pairs for english}.
\newblock \emph{TACL}.

\bibitem[{Warstadt et~al.(2019)Warstadt, Singh, and Bowman}]{CoLA}
Alex Warstadt, Amanpreet Singh, and Samuel~R. Bowman. 2019.
\newblock \href {https://transacl.org/ojs/index.php/tacl/article/view/1710}
  {Neural network acceptability judgments}.
\newblock \emph{TACL}.

\bibitem[{Williams et~al.(2018)Williams, Nangia, and Bowman}]{mnli-dataset}
Adina Williams, Nikita Nangia, and Samuel~R. Bowman. 2018.
\newblock \href {https://doi.org/10.18653/v1/n18-1101} {A broad-coverage
  challenge corpus for sentence understanding through inference}.
\newblock In \emph{Proc.\ of NAACL}.

\bibitem[{Zellers et~al.(2019)Zellers, Holtzman, Rashkin, Bisk, Farhadi,
  Roesner, and Choi}]{realnews-dataset}
Rowan Zellers, Ari Holtzman, Hannah Rashkin, Yonatan Bisk, Ali Farhadi,
  Franziska Roesner, and Yejin Choi. 2019.
\newblock \href
  {http://papers.nips.cc/paper/9106-defending-against-neural-fake-news}
  {Defending against neural fake news}.
\newblock In \emph{Proc.\ of NeurIPS}.

\bibitem[{Zhang et~al.(2018)Zhang, Liu, Liu, Gao, Duh, and Durme}]{ReCoRD}
Sheng Zhang, Xiaodong Liu, Jingjing Liu, Jianfeng Gao, Kevin Duh, and
  Benjamin~Van Durme. 2018.
\newblock \href {http://arxiv.org/abs/1810.12885} {{ReCoRD}: Bridging the gap
  between human and machine commonsense reading comprehension}.

\bibitem[{Zhang et~al.(2020)Zhang, Warstadt, Li, and Bowman}]{idea-crash}
Yian Zhang, Alex Warstadt, Haau{-}Sing Li, and Samuel~R. Bowman. 2020.
\newblock \href {https://arxiv.org/abs/2011.04946} {When do you need billions
  of words of pretraining data?}

\bibitem[{Zhou et~al.(2020)Zhou, Zhang, Cui, and Huang}]{cats-probe}
Xuhui Zhou, Yue Zhang, Leyang Cui, and Dandan Huang. 2020.
\newblock \href {https://aaai.org/ojs/index.php/AAAI/article/view/6523}
  {Evaluating commonsense in pre-trained language models}.
\newblock In \emph{Proc.\ of AAAI}.

\bibitem[{Zhu et~al.(2015)Zhu, Kiros, Zemel, Salakhutdinov, Urtasun, Torralba,
  and Fidler}]{bookcorpus-dataset}
Yukun Zhu, Ryan Kiros, Richard~S. Zemel, Ruslan Salakhutdinov, Raquel Urtasun,
  Antonio Torralba, and Sanja Fidler. 2015.
\newblock \href {https://doi.org/10.1109/ICCV.2015.11} {Aligning books and
  movies: Towards story-like visual explanations by watching movies and reading
  books}.
\newblock In \emph{Proc.\ of ICCV}.

\end{thebibliography}
